%% file: arxiv.tex
\PassOptionsToPackage{dvipsnames,table}{xcolor}
\documentclass[acmtog]{acmart}

\usepackage{graphicx}
\usepackage{booktabs}   
\usepackage[normalem]{ulem}

\usepackage{xspace}
\newcommand{\ours}{\textsc{LiveSVG}\xspace}
\newcommand{\ourhard}{ChallengeSVG\xspace}

\usepackage{color}

\makeatletter
\DeclareRobustCommand\onedot{\futurelet\@let@token\@onedot}
\def\@onedot{\ifx\@let@token.\else.\null\fi\xspace}
\def\eg{\emph{e.g}\onedot}

\makeatother

\usepackage[capitalize]{cleveref}

\usepackage[table, dvipsnames]{xcolor}
\usepackage{pifont}

\newcommand{\cmark}{{\color{ForestGreen}\ding{52}}}%
\newcommand{\mmark}{{\color{Dandelion}\ding{51}}}%
\newcommand{\xmark}{{\color{BrickRed}\ding{56}}}%
\setcopyright{none}
\renewcommand\footnotetextcopyrightpermission[1]{}
\renewcommand{\keywords}[1]{}
\renewcommand\footnotetextauthorsaddresses[1]{}
\AtBeginDocument{%
  \fancypagestyle{firstpagestyle}{%
    \fancyhf{}%
  }%
  \fancypagestyle{standardpagestyle}{%
    \fancyhf{}%
  }%
  \pagestyle{standardpagestyle}%
}
\begin{document}

\begin{teaserfigure}
\centering
  \includegraphics[width=0.95\linewidth]{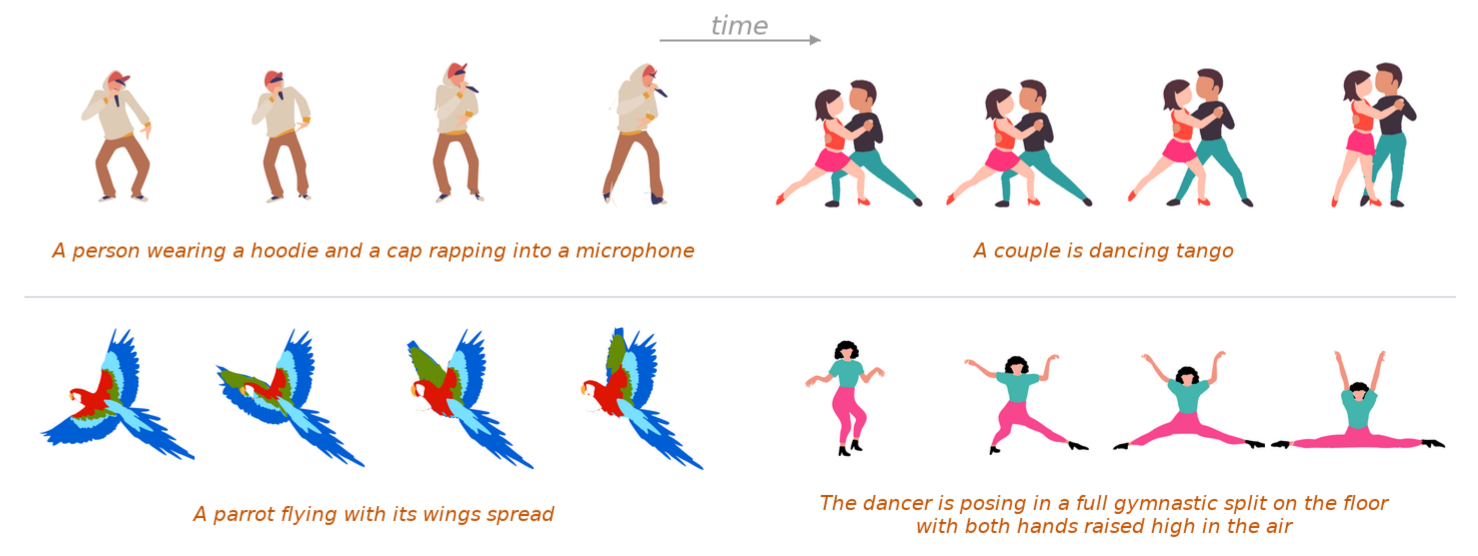}
  \caption{\ours animates a static SVG from a motion prompt in a zero-shot setting. It first generates a previewable target video with an image-to-video model, then fits the original vector geometry directly to the target sequence. The leftmost image in each example is the input SVG, followed by three frames from the resulting editable SVG animation. Project page: \href{https://levymsn.github.io/LiveSVG/}{\textcolor{orange}{https://levymsn.github.io/LiveSVG}} .}
  \Description{Rows of key frames showing static SVG inputs at the left and subsequent frames from the corresponding LiveSVG animated outputs, illustrating articulated motion across the sequence while preserving an editable vector representation.}
  \label{fig:teaser}
\end{teaserfigure}

\author{Matan Levy}
\affiliation{%
  \institution{Google}
  \country{Israel}}
\affiliation{%
  \institution{The Hebrew University of Jerusalem}
  \country{Israel}}

\author{Ran Margolin}
\affiliation{%
  \institution{Google}
  \country{Israel}}

\author{Bar Cavia}
\affiliation{%
  \institution{Google}
  \country{Israel}}

\author{Dvir Samuel}
\affiliation{%
  \institution{Bar-Ilan University}
  \country{Israel}}

\author{Yael Pritch}
\affiliation{%
  \institution{Google}
  \country{Israel}}

\author{Shmuel Peleg}
\affiliation{%
  \institution{The Hebrew University of Jerusalem}
  \country{Israel}}

\author{Alex Rav Acha}
\affiliation{%
  \institution{Google}
  \country{Israel}}

\author{Ariel Shamir}
\affiliation{%
  \institution{Google}
  \country{Israel}}
\affiliation{%
  \institution{Reichman University}
  \country{Israel}}

\author{Dani Lischinski}
\affiliation{%
  \institution{Google}
  \country{Israel}}
\affiliation{%
  \institution{The Hebrew University of Jerusalem}
  \country{Israel}}

\title{LiveSVG: Zero-Shot SVG Animation via Video Generation}




\input{sec/abstract}

\maketitle

\input{sec/introduction}
\input{sec/related_work}

\input{sec/method}
\input{sec/evaluation}

\input{sec/summary}

\bibliographystyle{ACM-Reference-Format}
\bibliography{sample-base}
\clearpage
\input{sec/last_figures}

\clearpage
{\Huge Supplementary Material}
\\
\appendix
\noindent
\input{sec/appendix}

\end{document}

%% file: sec/abstract.tex
\begin{abstract}
We introduce \ours, a zero-shot approach for generating Scalable Vector Graphics (SVG) animations using video diffusion models. Current SVG animation methods struggle with complex motions: LLM-based code synthesis fails to express fine, non-rigid B\'ezier deformations, while Score Distillation Sampling (SDS) provides noisy gradients and often requires category-specific priors like skeletons. In contrast, \ours fits vector geometry directly to an explicitly generated target video. Given an input SVG image and a motion prompt, we generate a previewable target video using a frozen image-to-video model, then fit the original SVG to this video via differentiable rendering. Our fitting stage is skeleton-free, utilizing a dual-level motion representation that combines per-group homographies for coarse articulation with per-path B\'ezier control-point offsets for local deformations. To resolve color-induced correspondence ambiguities during pixel-wise fitting, we introduce a novel sphere-packing recolorization strategy. We also present \textbf{\ourhard}, a benchmark of complex, multi-object scenes that exposes the limitations of prior work. Evaluations demonstrate that \ours significantly outperforms existing methods on both AniClipart and \ourhard, establishing direct reference-video fitting as a practical, robust route to prompt-aligned and fully editable vector animation.
\end{abstract}

%% file: sec/introduction.tex
\section{Introduction}
\label{sec:introduction}

Scalable Vector Graphics (SVGs)~\cite{w3c01svg} are a widely used format for 2D graphics such as icons, logos, and web illustrations. Unlike raster images, SVGs use explicit geometric primitives, such as B\'ezier curves and appearance attributes, yielding assets that are resolution-independent and directly editable. These properties make SVGs especially useful for reusable graphics and downstream design workflows. Although the SVG standard natively supports animation through time-varying attributes, generating high-quality animated SVGs remains significantly more difficult than creating static vector graphics.

Organic vector animation requires smoothly controlling the coherent deformation of many interconnected paths. Recent breakthroughs in static SVG generation~\cite{li2020diffvg,xing2024llm4svg,vector_fusion,star_vector,omniSVG} relied heavily on large datasets of paired raster images and vector graphics~\cite{xing2024llm4svg}. Unfortunately, comparable data for vector animation remains scarce~\cite{xiong2025internsvg,yang2026omnilottie,liang2026vanim}. Existing datasets predominantly feature simple UI elements or
procedural motion, failing to capture the complex, non-rigid deformations required for articulated subjects~\cite{liang2026vanim}. Consequently, large-scale supervised training for open-domain SVG animation remains impractical.

This data bottleneck has motivated zero-shot approaches that leverage the priors of large foundation models, broadly following two paradigms. The first uses LLMs or VLMs to generate SVG animation code directly~\cite{tseng2025keyframerempoweringanimationdesign,park2025decomate,lee2025vectorprism}. While frameworks like Vector Prism~\cite{lee2025vectorprism} successfully generate coherent motion programs, they struggle with fine-grained geometric deformations. Describing smooth, non-rigid curve motion at the coordinate level quickly exhausts LLM context limits, as SVG complexity increases. This forces code-generation methods to rely on coarse, rigid transformations that severely limit expressive animation~\cite{lee2025vectorprism}.

The second paradigm uses video diffusion models as stochastic motion priors via Video Score Distillation Sampling (SDS)~\cite{sds_loss,gal2023livesketch,huang2024aniclipart,khandelwal2025flexiclip,linrbridge2025}. However, SDS-based optimization suffers from several critical bottlenecks. The target motion remains implicit and noisy throughout the optimization loop, preventing users from previewing or editing the animation before the costly fitting process concludes. Furthermore, these pipelines frequently rely on category-specific scaffolds, such as skeletons or pose estimators, which restrict their applicability to humans or animals and significantly increase runtime.

In this work, we propose a fundamentally different paradigm: \emph{direct target-video fitting}. We entirely decouple animation generation from SVG optimization. Instead of using a video model as an in-the-loop stochastic prior, we first generate a concrete, previewable target video. This provides a major practical advantage: users can inspect, reject, or edit the target animation before optimization begins, and can easily integrate advanced closed-source generators like Veo~\cite{brooks2024veo}. The subsequent SVG fitting then becomes a well-defined reconstruction problem against a fixed target sequence. To make this pixel-wise fitting robust, we identify and solve a critical correspondence ambiguity: there can be confusion during optimization when different paths possess similar colors. We resolve this by temporarily recoloring paths using maximally separated RGB values derived from best-known sphere-packing solutions~\cite{sphere_packing_survey,packomania}, guaranteeing clean, unambiguous gradient flow.

We instantiate this approach in \ours, a skeleton-free, zero-shot framework that animates open-domain SVGs. Given a static input, \ours uses an LLM to partition elements into semantic groups, applies our sphere-packing recolorization, and generates a target animation using an image-to-video model. We then optimize the SVG via a differentiable renderer~\cite{li2020diffvg} using a dual-level motion representation. We learn per-group homographies to capture coarse articulation alongside per-path B\'ezier control-point offsets for fine, non-rigid deformations. As summarized in \Cref{tab:comparison}, this pipeline addresses the critical capability gaps of existing methods.

We evaluate \ours on the standard AniClipart benchmark~\cite{huang2024aniclipart} and further introduce \ourhard, a demanding new dataset of 35 complex, multi-object SVGs curated from SVGX-Core-250k~\cite{xing2024llm4svg}. Featuring layered occlusions and background-rich scenes, \ourhard exposes the brittleness of existing code-generation and SDS-based pipelines. Human preference studies show that \ours achieves state-of-the-art performance, with pairwise comparisons on both AniClipart and our challenging \ourhard benchmark. Furthermore, explicit target-video fitting proves highly efficient, requiring only a fraction of the runtime and GPU memory demanded by prior optimization-based baselines.

\begin{figure}[t]
  \centering
  \includegraphics[width=\linewidth]{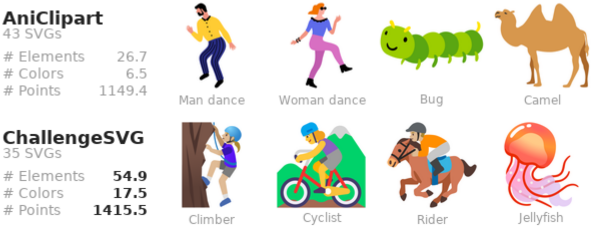}
  \caption{Representative images and structural statistics for AniClipart and \ourhard. The top row shows four AniClipart examples, while the bottom row shows four \ourhard examples. Left panels report mean dataset statistics. Most \ourhard images include multiple objects.}
  \Description{Two-row comparison figure. The top row shows four representative AniClipart SVG
  inputs with mean dataset statistics on the left. The bottom row shows four representative
  \ourhard SVG inputs with mean dataset statistics on the left, illustrating more complex
  multi-object scenes and denser vector structure.}
  \label{fig:svgx_hard_samples}
\end{figure}

\input{tables/svg_comparison}

\noindent In summary, our main contributions are:

\begin{itemize}
\item \textbf{\ours:} A novel zero-shot approach for animating static SVGs that decouples motion generation from vector fitting, optimizing the original geometry directly against an explicitly generated, previewable target video.
\item \textbf{Path-level recolorization:} A novel optimization strategy that utilizes the SVG structure to temporarily assign maximally separated RGB values to each path (region). Maximizing inter-path color distance helps eliminate correspondence ambiguities during pixel-space fitting.
\item \textbf{\ourhard benchmark:} A challenging new benchmark of complex, multi-object, background-rich SVGs designed to stress-test open-domain vector animation.
\item \textbf{State-of-the-art performance:} \ours achieves the highest human preference scores and best prompt-alignment metrics across both standard and challenging evaluation regimes.
\end{itemize}

%% file: tables/svg_comparison.tex
\begin{table}[t]
\caption{Comparison of prior SVG animation approaches and LiveSVG. We compare key practical capabilities for open-domain SVG animation: unconstrained motion, skeleton-free operation, and complex-scene handling. A half checkmark denotes partial support.}
\label{tab:comparison}
  \centering
  \small
  \setlength{\tabcolsep}{2.7pt}
  \resizebox{\columnwidth}{!}{%
  \begin{tabular}{lcccc}
    \toprule
    \multicolumn{1}{c}{Method} &
    \multicolumn{1}{c}{\shortstack{Unconstrained\\Motion}} &
    \multicolumn{1}{c}{\shortstack{Skeleton\\Free}} &
    \multicolumn{1}{c}{\shortstack{Open\\Domain}} &
    \multicolumn{1}{c}{\shortstack{Complex\\Scenes}} \\
    \midrule
    \multicolumn{5}{l}{\textit{LLM/VLM-based code synthesis}} \\[2pt]
    Vector Prism    & \xmark & \cmark & \mmark & \xmark \\
    \midrule
    \multicolumn{5}{l}{\textit{Optimization based on video SDS loss}} \\[2pt]
    LiveSketch      & \mmark & \cmark & \mmark & \xmark \\
    AniClipart      & \xmark & \xmark & \xmark & \xmark \\
    FlexiClip       & \xmark & \xmark & \xmark & \xmark \\
    LINR-Bridge     & \xmark & \cmark & \xmark & \xmark \\
    \midrule
    \multicolumn{5}{l}{\textit{Target (mp4) video based}} \\[2pt]
    \ours           & \cmark & \cmark & \cmark & \cmark \\
    \bottomrule
  \end{tabular}
  }
\end{table}

%% file: sec/related_work.tex
\section{Related Work}
\label{sec:related_work}

\subsection{Vector Graphics and Differentiable Rendering}
SVG is a code-based language for 2D graphics~\cite{w3c01svg} in which visual content is specified by explicit primitives and attributes rather than pixels. Although this makes SVGs compact, resolution-independent, and editable, it also means that geometry must be manipulated rigorously in order to animate it.

A central primitive in SVG is the longstanding B\'ezier curve \cite{bezier1967definition}, especially the cubic B\'ezier segment, which is defined by anchor points (endpoints) and control points that determine the tangent directions and the resulting curvature. Because complex outlines can be expressed as collections of such segments, B\'ezier curves are the standard expressive representation for vector paths in SVG.

Animation is also natively supported in SVG, through tags such as \texttt{<animate>}, which define variation over time. Broadly, motion can be achieved in two ways. One can explicitly vary element attributes over time, for example by changing B\'ezier control points across frames. Alternatively, one can apply time-varying transformations to an entire element or group, such as translation, rotation, or more general projective warps. The first mechanism captures non-rigid deformations, while the second gives a compact description of coarse global motion.

Recently, the differentiable vector renderer DiffVG~\cite{li2020diffvg} made it possible to optimize SVG geometry, color, and other parameters from raster-domain objectives while preserving the vector representation. However, DiffVG is fundamentally a static renderer: it produces a single image for a single set of SVG parameters. A natural extension to animation, as adapted by previous works \citet{gal2023livesketch,huang2024aniclipart,khandelwal2025flexiclip}, is to render $K$ images, one per time step, then render them as video frames. In this work, we improve this technique by simply defining \emph{<animate>} tags that directly encode the time-varying parameters of each path, resulting in one SVG animation file to render, rather than $K$ images that compose an MP4 video.

\subsection{SDS-based Vector Animation}

Because paired datasets of static and animated SVGs are scarce, most existing optimization-based methods tackle SVG animation in a zero-shot manner by borrowing motion priors from pretrained video diffusion models. These methods are based on Score Distillation Sampling (SDS) \cite{sds_loss}, which involve an open-source video model in the optimization process. In brief, SDS renders the current candidate video, perturbs it with diffusion noise at a random timestep, and uses the frozen video model's denoising prediction to define a gradient that nudges the rendered frames toward samples that better match the prompt and the prior. This makes it possible to optimize vector parameters without paired SVG-animation supervision. \citet{gal2023livesketch} introduced this paradigm for animating vector sketches. \citet{huang2024aniclipart} extended it to broader domains by introducing sparse motion keypoints and ARAP-based deformation. \citet{khandelwal2025flexiclip} further improves character animation with locality-preserving free-form deformations, while still relying on a skeleton-like motion scaffold. \citet{linrbridge2025} follows a different route: it first learns motion in a layered neural implicit representation under video diffusion guidance, then warps the original SVG to the neurally inferred animation to complete the process.

These methods demonstrate that video priors can animate vector content without dedicated training data, but they also inherit several limitations. First, the target motion remains implicit during optimization: SDS provides a stochastic gradient field rather than a fixed animation that can be inspected, accepted, or edited before fitting. Second, because the video prior stays in the optimization loop, supervision is noisy and computationally expensive compared with fitting to an explicit target sequence. Third, several methods introduce additional structural assumptions or intermediate representations, such as sketch-oriented inputs, keypoints, skeletons, ARAP-style deformation, or layered neural implicits. In contrast, our method is skeleton-free, keeps the source SVG as the object being optimized, and fits it to a concrete target video that is previewable before optimization begins.

\subsection{LLM/VLM-based SVG Animation}
Large language models and vision-language models have recently been explored for a wide range of SVG tasks, including SVG understanding, captioning, editing, and text/image-to-SVG generation~\cite{vector_fusion,star_vector,omniSVG,xiong2025internsvg}. Most of this literature focuses on static SVGs, where the model reasons over SVG code, raster renderings, or both, and then predicts or edits a vector program. Only very recent works have begun to address SVG animation directly~\cite{tseng2025keyframerempoweringanimationdesign,park2025decomate,lee2025vectorprism,xiong2025internsvg}. Vector Prism~\cite{lee2025vectorprism} tackles animation more directly in a zero-shot pipeline: it uses VLM-based semantic parsing to recover meaningful groups in the input SVG, uses an LLM/VLM-driven planner to decide how those groups should move, and then generates CSS/HTML-based animation code so that semantically related elements are animated coherently.

\subsection{Supervised Vector Animation Datasets}
To support such LLM-based approaches, recent large-scale vector datasets, such as SAgoge~\cite{xiong2025internsvg}, MMLottie-2M~\cite{yang2026omnilottie}, and SVGAnim-134k~\cite{liang2026vanim}, have been introduced to train LLMs for supervised code generation. However, they are fundamentally ill-suited for open-domain SVG fitting. These datasets predominantly feature simple UI icons or rigid procedural transforms. Consequently, they bias models toward simple CSS/SMIL animations and fail to capture the complex, non-rigid B\'ezier deformations required for articulated motion~\cite{liang2026vanim}. Furthermore, converting Lottie outputs back to standard SVGs often flattens native geometry, severely hindering downstream editability.

Despite these advancements, representing motion purely as generated program text (a paradigm reinforced by these supervised datasets) imposes a severe limitation on animation complexity. By relying on CSS or SMIL formats, these methods are inherently restricted to rigid, affine motions~\cite{liang2026vanim}. Consequently, they fail to capture the highly articulated, non-rigid deformations necessary for organic movement, such as a character bending its joints or shapes morphing fluidly. True free-form deformation requires direct, continuous manipulation of the underlying B\'ezier path geometry, a task that lies beyond the representational capacity of discrete CSS commands and leads to severe coordinate hallucinations when attempted by LLMs~\cite{liang2026vanim}. In short, generating a discrete program that merely \emph{describes} rigid motion, instead of fitting the continuous curves that realize it, makes fine path-level geometric control indirect and brittle.

Our approach addresses a fundamentally different regime. We bypass discrete code synthesis entirely and do not require simplified SVG animation training data. Instead, we optimize the original SVG geometry directly against an explicit reference video, allowing free-form path deformations. This provides direct, continuous curve-level supervision and a previewable motion target, enabling the complex, free-form B\'ezier movement that code-generation models have yet to achieve.

%% file: sec/method.tex
\section{Method}
\label{sec:method}

\subsection{Overview}

\begin{figure*}[t]
\centering
  \includegraphics[width=\linewidth]{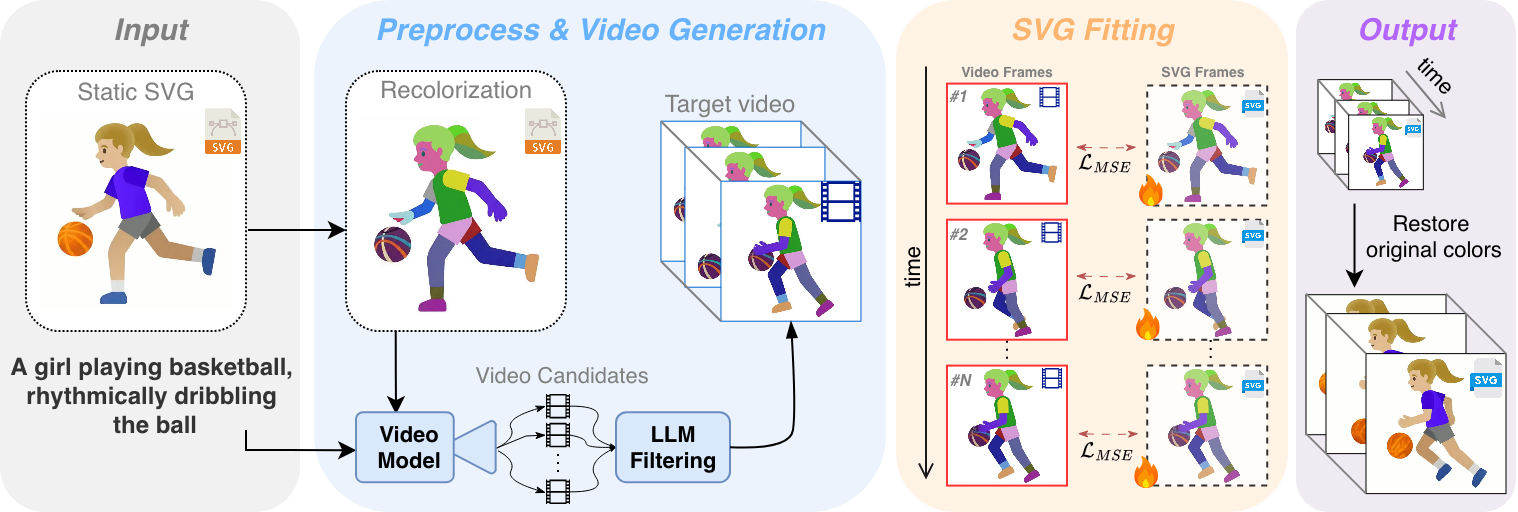}
  \caption{Overview of \ours.
  We first preprocess the input SVG with semantic grouping and sphere-packing recolorization,
  then generate candidate videos from the recolorized SVG and filter them to select the best target video.
  Finally, we fit each target frame back to the SVG by optimizing per-group homographies and
  per-path B\'ezier control-point deformations.}
  \Description{Pipeline diagram for \ours. An input SVG is preprocessed by grouping elements
  and applying sphere-packing recolorization. The recolorized SVG is used to generate multiple
  candidate videos, which are filtered to choose one target video. Each target frame is
  then fitted to the SVG using per-group homographies and per-path deformations.}
  \label{fig:pipeline}
\end{figure*}

Given a static SVG $S$, our goal is to produce an animated SVG $\hat{S}$ that preserves the
original vector structure while exhibiting plausible motion.
\ours decomposes this problem into two main stages, depicted in \Cref{fig:pipeline}.
First, a preprocessing stage groups SVG elements semantically, recolors paths with an optimization-friendly palette, and generates a target video.
Second, a differentiable optimization stage fits the SVG motion directly to the target video.
This separation between motion specification and vector fitting is deliberate: the target
video can be generated using any video model, whether open-source or accessed through a closed API. It can also be previewed, regenerated, or externally edited before optimization, while the
downstream SVG fitting objective remains unchanged. 
Full implementation details are available in the supplementary material.

\subsection{Stage 1: Preprocessing}

Given an input SVG, we perform three preprocessing phases before optimization.
First, we identify semantic groups that should be allowed to move together.
Next, we convert the SVG into an optimization-friendly colored representation that yields
cleaner gradients under pixel-space fitting, preventing identically colored paths from interfering with one another. Finally, we estimate a new layer order that is more consistent with the target motion, for edge cases in which different elements
occlude each other in the video.

\subsubsection{Semantic Grouping and Layer Ordering}

An input SVG typically contains many drawable elements, such as separate paths for facial parts,
limbs, clothing, or accessories. Animating all of them independently is unnecessarily complex for coarse motion, since many parts should move together as a single semantic unit. For example, when a character nods, the eyes, nose, mouth, hair, and head outline should all undergo the same global motion rather than be optimized as unrelated objects.

We therefore partition the SVG into semantic groups and later assign each group a shared learned global
transform. To obtain these groups, we use an LLM in a zero-shot manner. Specifically, we prompt the model to insert flat, non-nested \texttt{<g>} tags that group parts expected to move together.

The role of semantic grouping in \ours is important but not absolute:
it provides the model with a mechanism for coherent group-level motion, but the final animation is
not restricted to grouped rigid motion only. Since we later allow each path to deviate locally
through learned control-point offsets, moderate grouping mistakes are often tolerable in practice: the grouping stage supplies a useful motion prior, not a hard structural bottleneck.

For layer ordering, SVG rendering follows a painter's algorithm: elements appearing later in the file occlude earlier ones. Before fitting, we optionally estimate a better layer order from the target video:
We utilize SAM2~\cite{ravi2024sam2} to compare how much exclusive area each group preserves in regions where two groups compete. Intuitively, the foreground group keeps a more stable visible support, while the occluded group loses area when the two overlap. The resulting pairwise preferences are converted into a global layer order, with conservative fallbacks when the evidence is ambiguous.

\subsubsection{Sphere-packing Recolorization}

In this phase, we convert the SVG into an optimization-friendly representation for fitting the target video. Our differentiable objective matches rendered SVG frames to target frames using pixel-space
Mean Squared Error (MSE). By itself, this is prone to a serious ambiguity: if two distinct elements have similar colors, the optimizer can reduce loss by moving one element toward the other, even when the geometry is wrong, because the gradients only observe local color discrepancies and not semantic identity. For example, if a character's arm and leg have similar colors, crossings between them can become hard to resolve under a direct MSE objective.

To avoid such failure modes, we recolor the grouped SVG at the path level so that each
path receives a distinct color that is as far as possible from the others in RGB space.
Suppose the grouped SVG contains $P$ recolorable paths. We use a best-known solution to the sphere-packing-in-a-cube problem in $[0,1]^3$ \cite{sphere_packing_survey}, published by Packomania~\cite{packomania}, and choose the $P$ sphere centers as RGB colors. If the packing radius is $r$, then every pair of chosen colors is separated by at least $2r$. This gives a principled palette in which all recolored paths are highly distinguishable to the loss.

The recolorized SVG is used only during target generation and optimization. The final exported animation is written back in the original colors. In this way, sphere-packing recolorization improves optimization stability without changing the editable SVG content that the user ultimately receives.

\subsection{Target Video Generation and Filtering}

Given the optimization-friendly SVG, we generate a target animation with a frozen
image-to-video model. We first render the recolorized SVG to an image and provide it, together with the desired animation prompt, to a diffusion image-to-video model. Specific model choices are deferred to the implementation details.

Our constructed prompt explicitly asks for motions that are compatible with SVG fitting. In particular, we request \emph{flat 2D vector-style animation}, \emph{solid block colors}, \emph{no new details}, \emph{preservation of the colors}, etc. These constraints matter because our approach assumes fixed-color SVG elements and a fixed set of vector primitives. Color drift may produce false MSE gradients, 3D out-of-plane motion may reveal geometry that cannot be represented in the SVG, and invented elements may not necessarily be represented by the source vector structure.

Despite recent progress, image-to-video models are still prone to failure cases such as implausible motion, unintended 3D rotation, color changes over time, or the invention/removal of elements.
We therefore generate $10-20$ candidate videos for each SVG using different random seeds and automatically score them with an LLM. The first-phase scorer evaluates criteria tailored to downstream SVG fitting, including whether the motion remains strictly 2D, whether only the original elements are present, whether physical continuity is preserved, whether flat colors remain constant, and whether the video is suitable as a target for fixed-color MSE tracking. We then comparatively rerank the top candidates with the same LLM in a second stage and select the highest-ranked video. The details of this scoring and reranking procedure are deferred to the supplementary material. The top-ranked video is treated as our target video.

\subsection{Stage 2: Target Fitting Optimization}
In this phase, we fit the SVG animation to the selected target video.

\subsubsection{Vector Motion Parameterization}
\ours represents motion at two complementary levels: a shared transform per semantic group for
coarse motion, and local control-point offsets per path for non-rigid deformations.

\paragraph{Per-group global motion.}
Each semantic group is assigned a learnable $8$-DOF homography per frame.
Let $H_{g,k}$ denote the transform for group $g$ at keyframe $k$, with the first keyframe anchored
at the identity transform.

\paragraph{Per-path local deformation.}
Group-level motion alone cannot capture articulated or elastic changes.
We therefore assign keyframe-dependent control-point offsets to each path. These offsets are stored as learned
per-keyframe embeddings. For a path with canonical control points $p_i$, the deformed point at keyframe $k$ is
$\tilde{p}_{i,k} = p_i + \Delta_{i,k}$,
where $\Delta_{i,k}$ is the learned local displacement.

This two-level parameterization makes local and global motion easier to control.
Homographies capture coherent part motion such as translation, rotation, and scale, while path offsets capture free-form, non-rigid phenomena such as bending,
squash-and-stretch, or any local shape-specific motion.

\subsubsection{Frame Fitting Optimization}

Given the target video, we optimize the SVG parameters so that rendered frames
match the extracted target frames.
The optimization acts directly on the vector representation and propagates gradients to the learned homographies and path offsets.

Our main data term $\mathcal{L}_{\mathrm{MSE}}$ is a blurred pixel-space Mean Squared Error (MSE) loss.
Let $R_k$ be the video's $k$-th target frame, $I_k$ be the $k$-th differentiably rendered SVG frame, and $\mathcal{G}$ be a Gaussian blur operator. Then:
\begin{equation}
\mathcal{L}_{\mathrm{MSE}} =
\frac{1}{K}\sum_{k=0}^{K-1}
\left\|\mathcal{G}(R_k) - \mathcal{G}(I_k)\right\|_2^2
\end{equation}

We further regularize the local path deformations using exponential spatial regularization, which encourages nearby control points on the same path to move coherently while allowing more distant points to deform independently.
For adjacent control-point pairs $(i,j)$, we use
\begin{equation}
\mathcal{L}_{\mathrm{spatial}} =
\frac{1}{K}\sum_{k=0}^{K-1}\sum_{(i,j)\in\mathcal{D}}\exp\!\left(-\left(\frac{\|p_i-p_j\|_2}{\sigma}\right)^2\right)
\left\|\Delta_{i,k}-\Delta_{j,k}\right\|_2^2
\end{equation}
where $\mathcal{D}$ contains adjacent control-point pairs from the path element, and $\sigma$ is a fixed fraction of the canvas width.
This regularizer suppresses noisy local motion while preserving the ability to model meaningful non-rigid deformation.

Two additional geometric regularizers stabilize the optimized path shapes.
First, we encourage $G^1$ continuity \cite{g1_continuity} at cubic B\'ezier joints,
for each shared anchor with an original angle difference of $\pm 10^{\circ}$.
This discourages the optimizer from explaining pixel errors by introducing sharp kinks at joints that originally were visually smooth.

Second, we use a foreground signed distance function (SDF) penalty.
For each target keyframe, we compute a foreground mask and its signed distance field
$D_k(x)$, where positive values indicate distance outside the foreground.
Let $q_{i,k}$ be a rendered control point after applying the local offset and group homography.
We penalize points that drift outside the foreground by more than a small margin $\tau$ with:
\begin{equation}
\mathcal{L}_{\mathrm{SDF}} =
\frac{1}{K}\sum_k \frac{1}{|\mathcal{P}|}\sum_i
\max(0, D_k(q_{i,k})-\tau)^2 .
\end{equation}
Here $\mathcal{P}$ is the set of optimized control points.
This term is not a silhouette loss; it is a geometric guardrail that keeps raw SVG control points
from escaping the object support while the MSE term fits the visible appearance.

The final loss we minimize is defined as:
\begin{equation}
\mathcal{L} =
\lambda_{\mathrm{mse}}\mathcal{L}_{\mathrm{MSE}} +
\lambda_{\mathrm{spatial}}\mathcal{L}_{\mathrm{spatial}} +
\lambda_{\mathrm{g1}}\mathcal{L}_{\mathrm{G1}} +
\lambda_{\mathrm{sdf}}\mathcal{L}_{\mathrm{SDF}} .
\end{equation}
\subsubsection{Progressive Optimization Over Frames}

Optimizing all frames independently from the static SVG initialization might be unreliable, as the reconstruction loss $\mathcal{L}_{\mathrm{MSE}}$ only provides useful gradients when the rendered element already overlaps its target, at least partially. If an object moves too far from its initial location, the optimizer may receive little or no signal about how to recover it. A simple example is a ball moving from the top of the image to the bottom: after several frames, initializing each frame directly from the rest pose can place the predicted ball so far from its target that the MSE objective provides a very weak alignment signal.

To avoid this problem, \ours uses progressive optimization over frames.
We first optimize the early active frames, then activate a new frame and initialize its
parameters from the previous one. Concretely, when keyframe $k+1$ becomes active, we copy the homography parameters and the local path-deformation parameters from keyframe $k$.
This guarantees a plausible initial overlap between consecutive frames and lets the optimizer
explain motion incrementally rather than solving the full sequence from scratch.

Overall, the progressive schedule is crucial: it turns a hard long-range alignment problem into a sequence of easier local updates.

\subsubsection{Tracking-based Keyframe Initialization}

Copying the previous keyframe provides a useful initialization, but it can still be too weak when a
new frame contains large motion or partial occlusion (\eg low-FPS video).
We therefore augment each progressive transition with point-tracking initialization.
When keyframe $k+1$ is activated, we sample on-curve B\'ezier anchor points from frames $\{0,\ldots,k\}$ and track them into the incoming target frame using TAPNext~\cite{doersch2024tapnext}. High-confidence points are used to predict the point offsets in $k+1$. Finally, we use a conservative fallback for unreliable tracked paths:
we compare the tracking-based initialization against the previously learned version of the path by measuring masked reconstruction error against the target frame $k+1$.
The best candidate is selected; this safety net prevents a single tracking failure from propagating through the rest of the progressive schedule, while still allowing tracking to provide strong overlap whenever it is reliable.

\subsection{Implementation Details}
We use Gemini~3.1~Pro~\cite{team2024gemini} for semantic grouping and candidate video filtering. For target video generation, we examine two open-source video models, LTX 2.3 \cite{hacohen2025ltxvideo} and WAN 2.2 \cite{wan2025}, and the proprietary closed-API model Veo~3.1~\cite{brooks2024veo}. We utilize DiffVG~\cite{li2020diffvg} for differentiable rasterization. We optimize at $256 \times 256$ resolution and export higher-resolution visualizations at 720 pixels. The default run uses 15 keyframes and 2000 optimization iterations. More details are available in the suppl. material.

%% file: sec/evaluation.tex
\section{Evaluation}
\label{sec:evaluation}
We evaluate \ours for SVG Animation using human preferences, automatic metrics, and resource measurements.

\subsection{Experimental Setup}

We evaluate \ours on AniClipart~\cite{huang2024aniclipart}, a 43-example benchmark spanning human actions, animals, and object-centric motion. We also introduce \textbf{\ourhard}, a 35-example benchmark collected from SVGX-Core-250k~\cite{xing2024llm4svg}, designed to stress multi-object scenes, non-empty backgrounds, dense path/color structure, and subjects not naturally described by a single human or animal skeleton. We compare against five available baselines: the optimization-based video-prior methods LiveSketch~\cite{gal2023livesketch}, AniClipart~\cite{huang2024aniclipart}, FlexiClip~\cite{khandelwal2025flexiclip}, and LINR-Bridge~\cite{linrbridge2025}, plus Vector Prism~\cite{lee2025vectorprism}, a VLM-based method that synthesizes SVG animation code.

\subsection{Qualitative Results}

\Cref{fig:qual_examples} shows the qualitative breadth of \ours across prompts and SVGs. The examples include complex geometry changes such as articulated boxing arms with depth-like foreshortening, surfer body rotation, wing and limb deformation, and object motion, while preserving the input structure.

\Cref{fig:qual_comp,fig:qual_comp_2} compare \ours with prior methods on large pose-change prompts. For ``a man sits on the floor'', \ours lowers the torso, extends the legs, and deforms the paths continuously into a sitting configuration. For the full-split dancer prompt, \ours transforms a standing character into a floor split with raised arms. In both cases, competing methods remain closer to the input pose or produce limited transform-centric limb motion, illustrating the difficulty of obtaining large deformation from skeleton templates, implicit SDS supervision, or direct code synthesis. Finally, \Cref{fig:diversity_comp} shows that our target-video fitting approach can produce diverse plausible animations for the same SVG and prompt, such as distinct rapping gestures and waving trajectories, while preserving the source structure.

\subsection{Human Preference Studies}
\label{sec:user_study}

We conduct a pairwise human preference study, evaluating our method against previous works. Workers on Amazon Mechanical Turk compare unlabeled animations generated from the same input SVG and prompt, asked for overall visual preference. Full survey details are reported in the suppl. material.

\begin{figure}[t]
  \centering
  \includegraphics[width=0.92\linewidth]{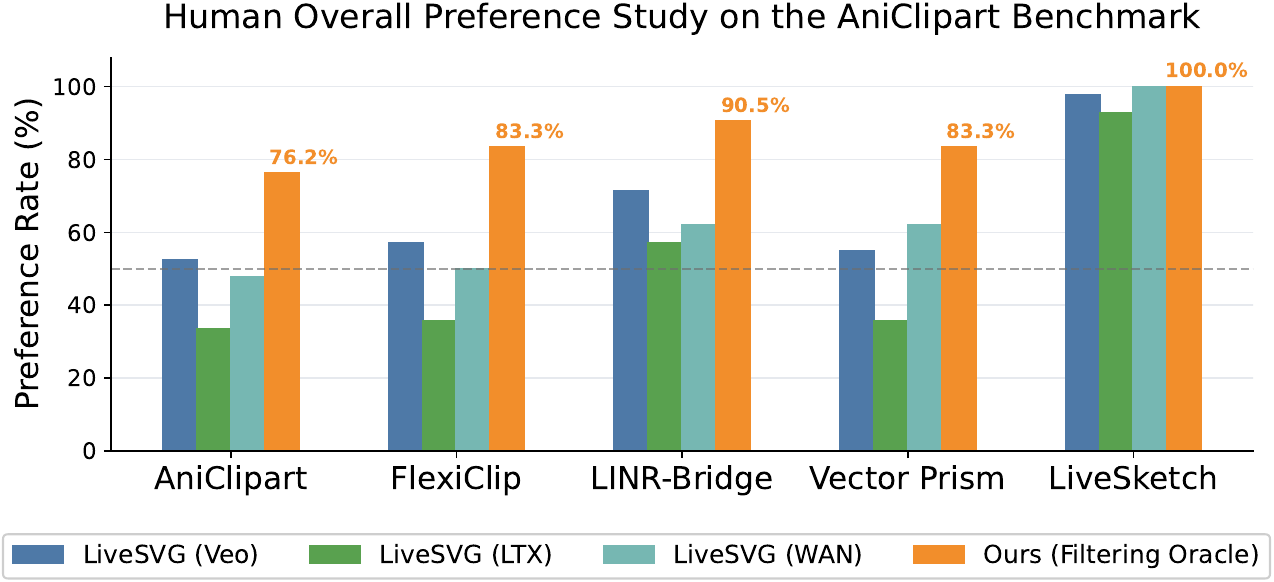}
  \caption{Human user study results on AniClipart. Each group reports pairwise overall preference scores against one baseline. \emph{\ours (Filtering Oracle)} selects, for each example, the best-of-three target video.}
  \Description{Grouped bar chart of human preference scores on AniClipart for LiveSVG with Veo, LTX, WAN, best-of-three aggregation, and the baseline complement against AniClipart, FlexiClip, LINR-Bridge, Vector Prism, and LiveSketch.}
  \label{fig:user_study_aniclipart_main}
\end{figure}

\begin{figure}[t]
  \centering
  \includegraphics[width=0.94\linewidth]{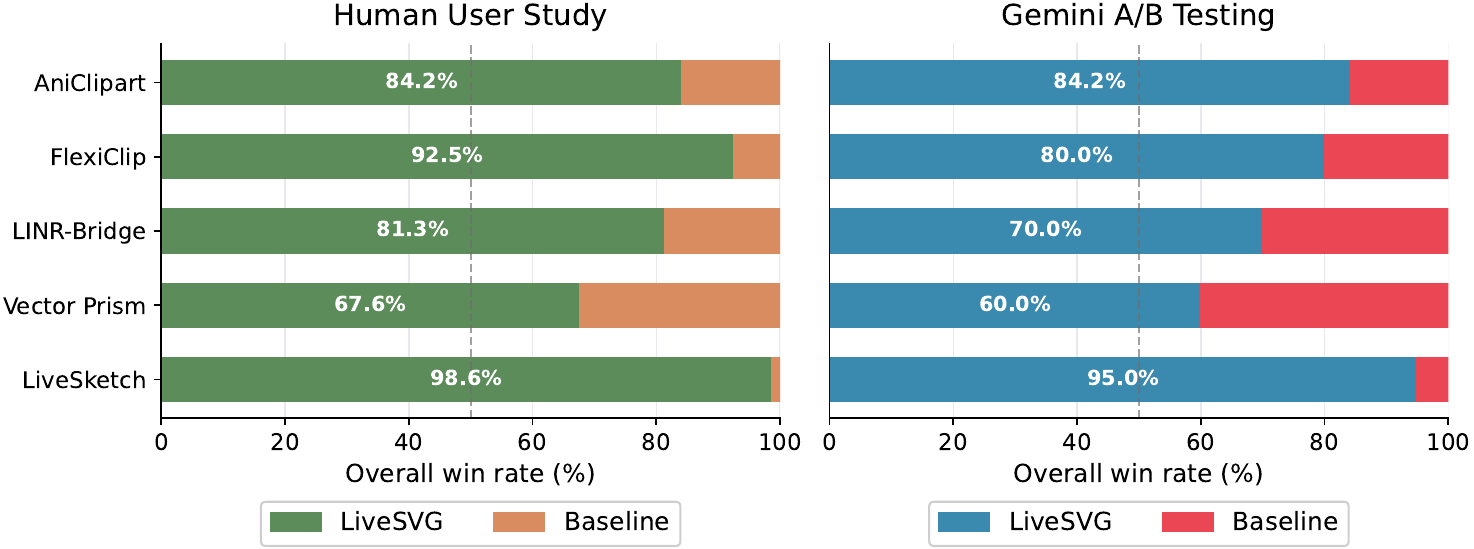}
  \caption{Overall preference win rates on \ourhard.
           Left: human user study results from Amazon Mechanical Turk.
           Right: automated A/B results from a Gemini-based judge.
           Both panels report pairwise overall win rates for \ours against each baseline.}
  \Description{Two horizontal stacked bar charts for \ourhard. The left chart shows human
  overall preference win rates for LiveSVG against AniClipart, FlexiClip, LINR-Bridge, Vector
  Prism, and LiveSketch. The right chart shows Gemini A/B overall win rates against the same
  baselines.}
  \label{fig:user_study_results}
\end{figure}

On the widely adopted AniClipart, we test three variants of \ours, each using a different video model: \emph{\ours (Veo~3.1)}~\cite{brooks2024veo}, \emph{\ours (LTX~2.3)}~\cite{hacohen2025ltxvideo}, and \emph{\ours (WAN~2.2)}~\cite{wan2025}. We also report a per-example \emph{\ours (Filtering Oracle)} result, as an Oracle selecting the best target video variant for each example. This reflects the practical workflow enabled by our approach: a user can inspect the previewable target video and choose the best target before SVG fitting. \ours wins \textbf{86.7\%} of human preferences on AniClipart (182/210) across the five baselines. The fixed variants win \textbf{66.7\%} for \emph{\ours (Veo~3.1)}, \textbf{64.3\%} for \emph{\ours (WAN~2.2)}, and \textbf{51.0\%} for \emph{\ours (LTX~2.3)}. In the per-baseline breakdown, \emph{\ours (Veo~3.1)} wins against all baselines, while \emph{\ours (WAN~2.2)} is competitive with AniClipart and FlexiClip. This sensitivity to the reference generator is expected in our plug-and-play design: as image-to-video models improve, the same SVG fitting stage can directly benefit from better target motion. \Cref{fig:user_study_aniclipart_main} reports the full variant breakdown.
On the \ourhard benchmark, \ours further wins \textbf{84.8\%} of human preferences (307/362), as shown in \Cref{fig:user_study_results}. This demonstrates the robustness of our approach across more challenging scenarios. As further validation, a Gemini-based A/B study shows the same trend, with \ours winning \textbf{64.6\%} of overall comparisons on AniClipart and \textbf{77.8\%} on \ourhard. Together, these results indicate that the gains persist beyond the simpler AniClipart setting.

\subsection{Quantitative Results}

\Cref{tab:quant} summarizes automatic metrics and resource usage on AniClipart. Protocol details and further results on \ourhard are provided in the suppl. material. DOVER~\cite{wu2023dover} measures overall perceptual video quality, diversity reports cross-seed variation for a fixed prompt, X-CLIP~\cite{ma2022xclip} measures prompt alignment, and time/VMem report the mean wall-clock time and GPU memory usage per SVG. The most significant differences are that \ours achieves the best prompt alignment score, the best appearance preservation among optimization-based methods, and the lowest GPU cost among optimization baselines (\textbf{5.2} minutes and \textbf{7.4 GB} per SVG). The LTX and WAN variants also give the strongest cross-seed diversity, supporting the benefit of decoupling motion generation from SVG fitting. 

\input{tables/metrics}

%% file: tables/metrics.tex
\begin{table}[t]
  \caption{Quantitative comparison on the AniClipart benchmark. For optimization-based methods, \textbf{bold} marks the best value and underlining marks the second best.}
  \label{tab:quant}
  \centering
  \small
  \setlength{\tabcolsep}{5pt}
  \resizebox{\columnwidth}{!}{%
\begin{tabular}{l ccccccc}
  \toprule
  Method & XCLIP$\uparrow$ & LPIPS$\downarrow$ & SSIM$\uparrow$ & DOVER$\uparrow$ & Diversity$\uparrow$ & Time$\downarrow$ & VMem$\downarrow$ \\
  \midrule
  No Animation & 0.211 & \textbf{0.000} & \textbf{1.000} & 0.444 & 0.000 & -- & -- \\
  \midrule
  \multicolumn{8}{l}{\textit{LLM/VLM-based}} \\[2pt]
  Vector Prism & 0.211 & 0.032 & 0.973 & 0.451 & 0.012 & 2.9m & -- \\
  \midrule
  \multicolumn{8}{l}{\textit{Optimization based on video SDS loss}} \\[2pt]
  LiveSketch & 0.206 & 0.153 & 0.910 & \textbf{0.496} & 0.026 & 35.9m & 35.1\tiny{GB} \\
  AniClipart & 0.214 & 0.104 & 0.937 & 0.427 & 0.020 & \underline{22.1}m & 27.9\tiny{GB} \\
  FlexiClip & 0.213 & \underline{0.092} & 0.938 & 0.431 & 0.053 & 57.4m & 28.1\tiny{GB} \\
  LINR-Bridge & \underline{0.215} & 0.174 & 0.925 & 0.433 & 0.016 & 79.5m & \underline{16.8}\tiny{GB} \\
  \midrule
  \multicolumn{8}{l}{\textit{Target (mp4) video based}} \\[2pt]
  \textbf{\ours (Veo~3.1)} & \textbf{0.216} & \textbf{0.087} & \textbf{0.942} & \underline{0.447} & 0.052 & \textbf{9.0}m & \textbf{7.1}\tiny{GB} \\
  \textbf{\ours (LTX~2.3)} & \underline{0.215} & 0.105 & \underline{0.940} & 0.445 & \textbf{0.099} & \textbf{9.0}m & \textbf{7.1}\tiny{GB} \\
  \textbf{\ours (WAN~2.2)} & 0.214 & 0.116 & 0.938 & 0.446 & \underline{0.063} & \textbf{9.0}m & \textbf{7.1}\tiny{GB} \\
  \bottomrule
\end{tabular}
  }
\end{table}

%% file: sec/summary.tex
\section{Summary}
\label{sec:summary}
In this paper, we presented \ours, a zero-shot approach for animating static SVGs with video models.

\paragraph{Limitations:} Our approach depends on the quality and SVG compatibility of the generated target video. Color drift, invented parts, or large occlusions can harm the optimized SVG and introduce artifacts. While our non-template optimization is flexible and outperforms prior baselines, it can still struggle in complicated scenes, especially when important regions are repeatedly occluded.

Despite these limitations, \ours achieves state-of-the-art results across AniClipart and \ourhard, showing stronger prompt alignment and appearance preservation than prior SVG-animation baselines. Beyond the quantitative gains, the target-video approach gives users a previewable intermediate result: they can inspect or regenerate the video before committing
to SVG optimization.

This design also opens the door to new editing workflows. Since \ours fits an SVG animation to a target video, future work could use video editing or video-control tools to edit the target first, then transfer the edited motion back to the vector domain. More broadly, we believe this work is
a step toward in-the-wild video-to-SVG animation, where rich motions from generated or captured videos can be converted into editable, compact vector animations.

%% file: sec/last_figures.tex
\begin{figure*}[ht]
  \centering
  \includegraphics[width=\linewidth]{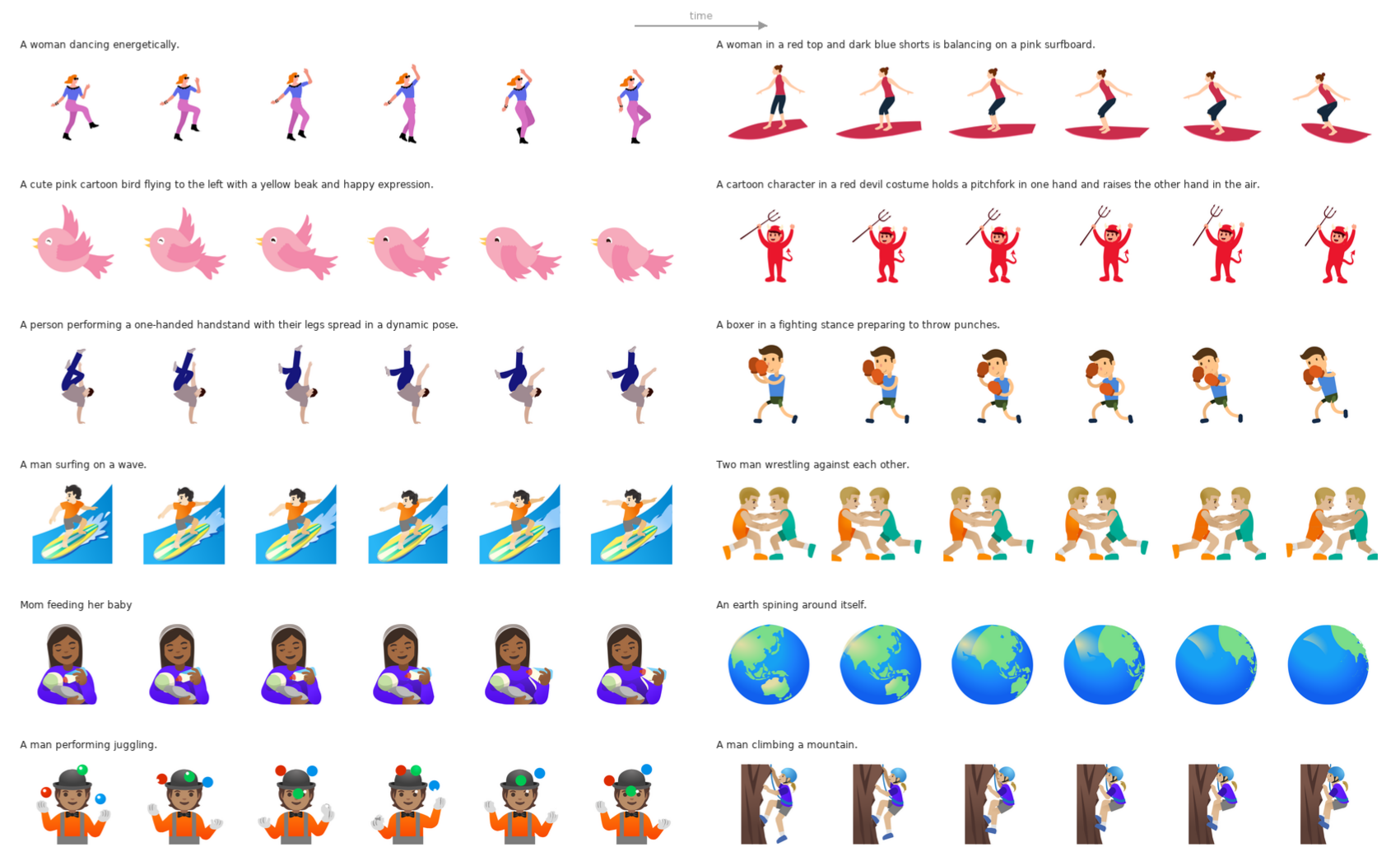}
  \caption{Qualitative breadth of \ours across diverse prompts and SVG structures.}
  \Description{Multiple LiveSVG outputs. Each example shows an input prompt and four animation
  frames with visible pose or shape changes, including dancing, surfing, flying, handstand, and
  martial-arts motions.}
  \label{fig:qual_examples}
\end{figure*}

\begin{figure*}[ht]
  \centering
  \includegraphics[width=0.95\linewidth]{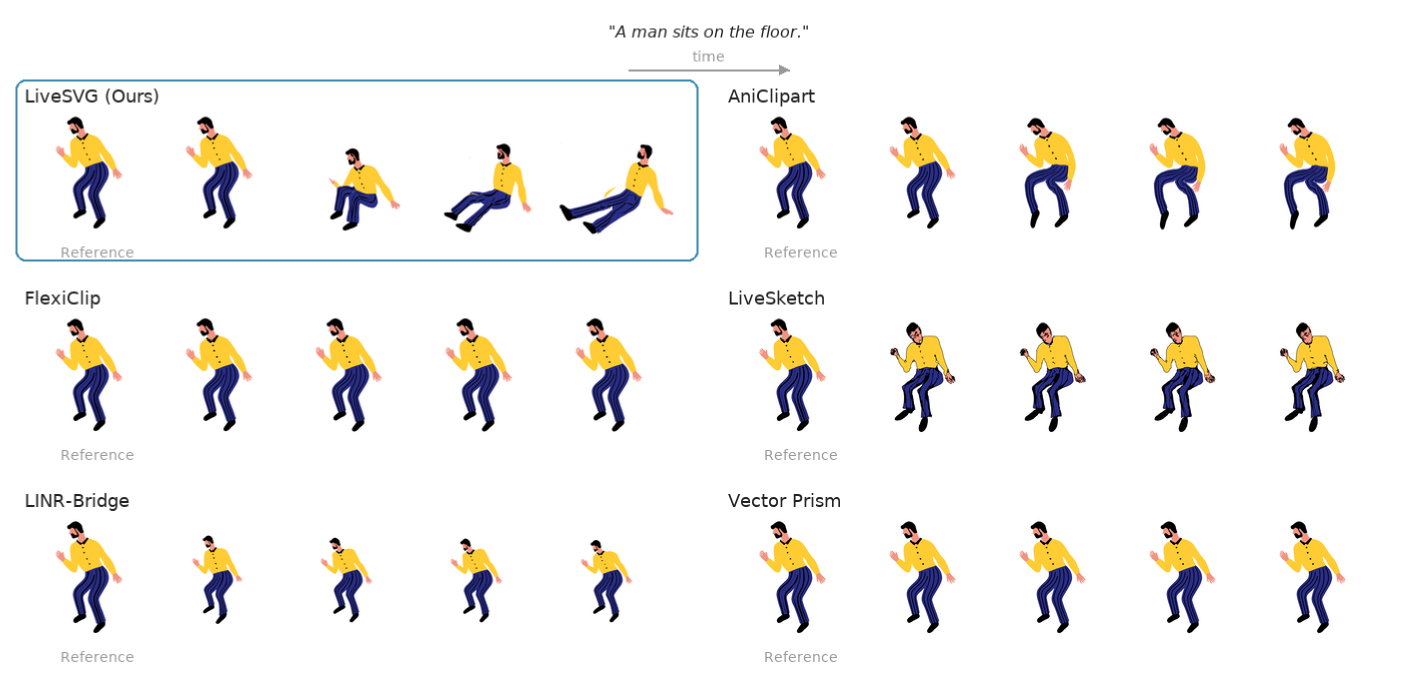}
  \caption{Qualitative comparison for the prompt ``A man sits on the floor''.
           \ours demonstrates a large topology-preserving pose change: the torso lowers and the legs extend into a sitting
           configuration. Competing methods remain closer to the standing input or produce a limited motion.}
  \Description{Side-by-side comparison of LiveSVG and baselines on a man-sitting prompt. LiveSVG
  changes the standing figure into a seated pose over time, while AniClipart, FlexiClip,
  LiveSketch, LINR-Bridge, and Vector Prism show more limited motion.}
  \label{fig:qual_comp}
\end{figure*}

\begin{figure*}[ht]
  \centering
  \includegraphics[width=0.95\linewidth]{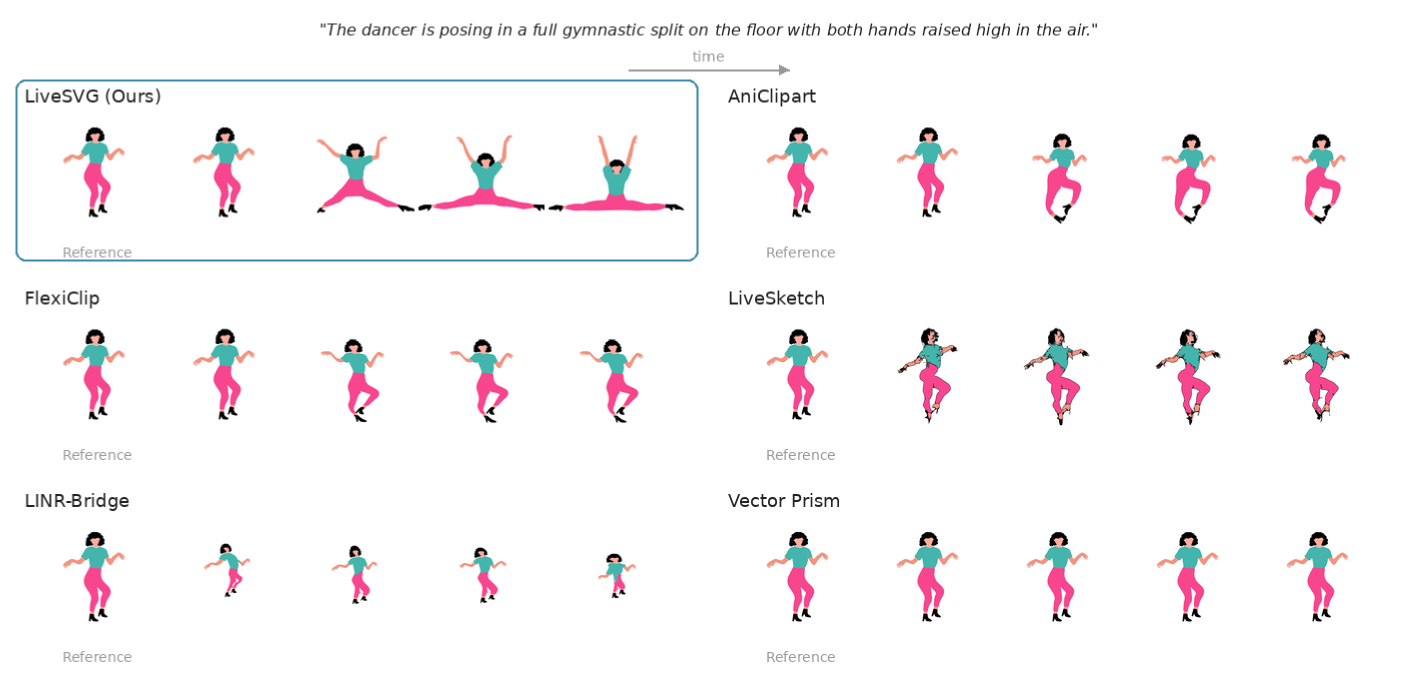}
  \caption{Qualitative comparison on a full-split dancer prompt. \ours lowers the dancer to the
           floor, extends both legs, and raises both arms, while competing methods mostly remain
           near the standing input or show limited limb motion.}
  \Description{LiveSVG produces a floor split with raised arms, while baselines remain mostly
  upright or show smaller motions.}
  \label{fig:qual_comp_2}
\end{figure*}

\begin{figure*}[ht]
  \centering
  \includegraphics[width=0.95\linewidth]{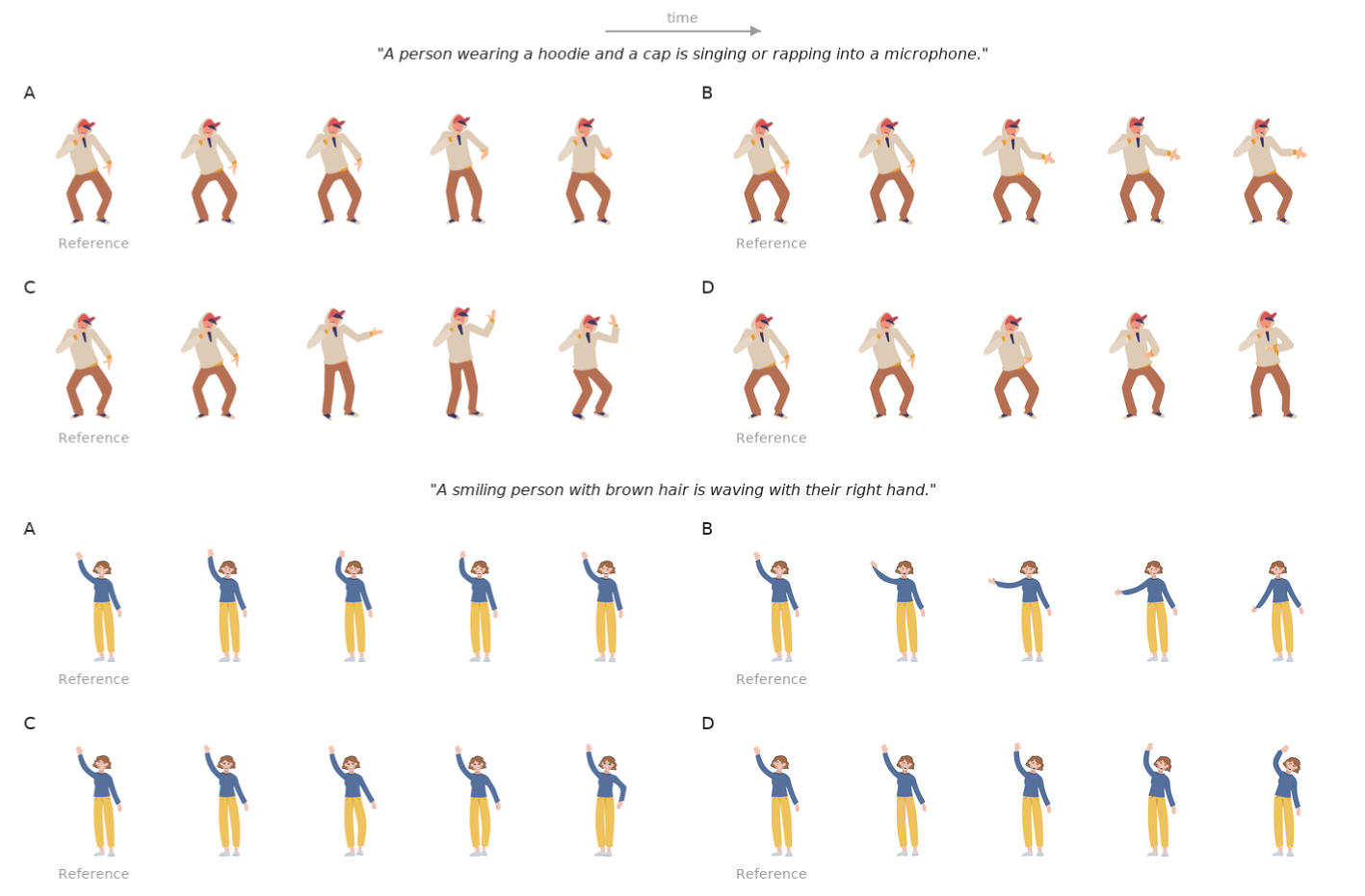}
  \caption{Seed diversity of \ours. Variants A--D preserve the SVG structure while producing
           distinct plausible motions for the same input and prompt.}
  \Description{Two LiveSVG examples with four seed variants each, showing different rapping gestures
  and arm-wave trajectories.}
  \label{fig:diversity_comp}
\end{figure*}

%% file: sec/appendix.tex
This supplementary material provides implementation details, evaluation protocols, and additional
quantitative results that complement the main paper. We avoid repeating the qualitative examples
and main human-study figures already included in the paper.

\section{Additional Results on \ourhard}

\paragraph{Benchmark structure statistics.}
To quantify the increased structural difficulty of \ourhard, we measure the static SVG
inputs used by the two benchmarks after the same grouping stage.
We count non-root SVG XML elements, distinct explicit fill/stroke/stop colors, and numeric
coordinate values in path \texttt{d} attributes.
\Cref{tab:benchmark_svg_stats} reports the resulting statistics.
\ourhard contains substantially richer input structure: compared with AniClipart, it has
2.1$\times$ more SVG elements on average, 2.7$\times$ more distinct colors on average, and a
higher median number of path-coordinate values.
These statistics support the intended role of \ourhard as a compact stress test for
multi-object, layered, background-rich SVGs.

\input{tables/benchmark_svg_stats}

\paragraph{Quantitative metrics and runtime.}
Table~\ref{tab:quant_svgx_hard} reports the same quantitative metrics used in the main
AniClipart evaluation, evaluated on \ourhard. This benchmark uses the \emph{\ours (WAN~2.2)}
variant for \ours, matching the human preference study configuration in
\Cref{sec:appendix_user_study}.
The same table also reports the per-sample runtime measured on five hard examples; the detailed
runtime breakdown is provided in \Cref{tab:appendix_runtime_svgx_hard}.
On this subset, \ours is the fastest GPU optimization method at 4.7 minutes per successful sample,
while several baselines slow substantially on complex SVGs.
This difference reflects the fact that methods such as Vector Prism and skeleton/template-heavy
optimization baselines expose latency that depends on serialized SVG element complexity or
structural setup, whereas \ours mainly optimizes against a fixed set of rendered keyframes rather
than asking a model to synthesize per-element animation code.

\input{tables/metrics_svgx_hard}

\input{tables/runtime_comparison_svgx_hard}

\section{Additional A/B Results}

\begin{figure*}[t]
  \centering
  \includegraphics[width=0.92\linewidth]{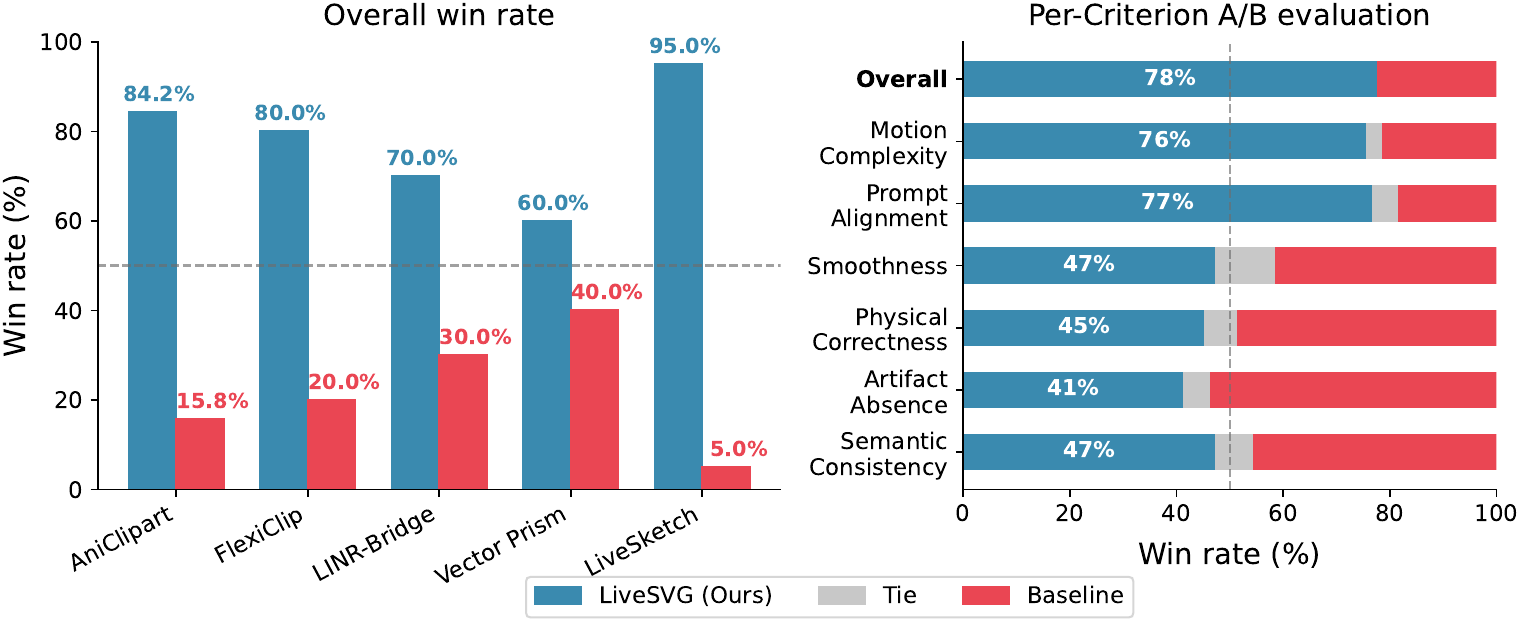}
  \caption{Automated A/B preference study results on \ourhard with a Gemini-based judge.
           Left: per-criterion outcomes.
           Right: overall win rate against each baseline.
           \ours wins 77.8\% of 99 overall comparisons.}
  \Description{Two charts summarizing the automated A/B study for LiveSVG on \ourhard. The
  left panel shows per-criterion win, tie, and baseline-win percentages. The right panel shows
  overall win rates against AniClipart, FlexiClip, LINR-Bridge, Vector Prism, and LiveSketch.}
  \label{fig:ab_results_svgx_hard}
\end{figure*}

\paragraph{\ourhard automated A/B results.}
Figure~\ref{fig:ab_results_svgx_hard} reports the automated A/B preference study on \ourhard.
The main paper reports the human preference results; this plot provides a complementary
criterion-level model-based signal.

\begin{figure*}[t]
  \centering
  \includegraphics[width=0.92\linewidth]{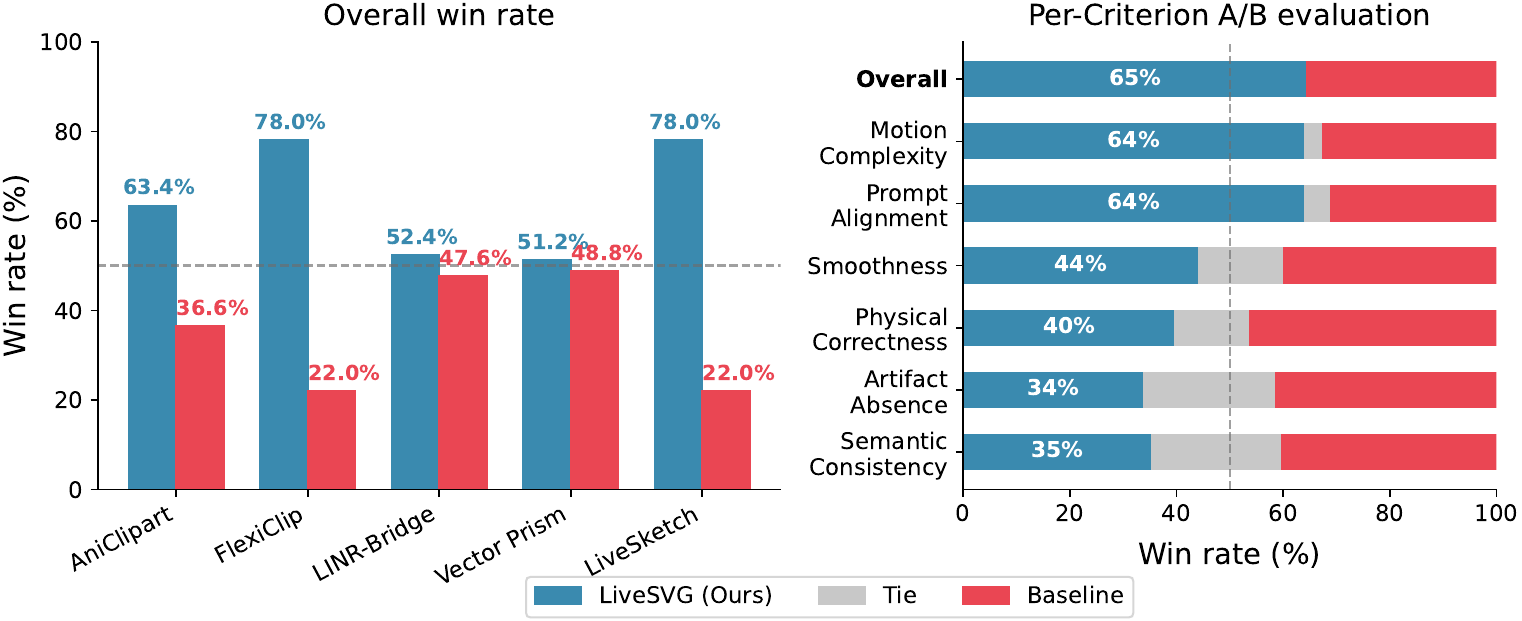}
  \caption{Automated A/B preference study results for the main AniClipart setting with a
           Gemini-based judge.
           Left: per-criterion outcomes.
           Right: overall win rate against each baseline.
           \ours wins 64.6\% of 206 overall comparisons.}
  \Description{Two charts summarizing the automated A/B study for LiveSVG on AniClipart. The left
  panel shows per-criterion win, tie, and baseline-win percentages. The right panel shows
  overall win rates against AniClipart, FlexiClip, LINR-Bridge, Vector Prism, and LiveSketch.}
  \label{fig:ab_results}
\end{figure*}

\begin{figure*}[t]
  \centering
  \includegraphics[width=0.92\linewidth]{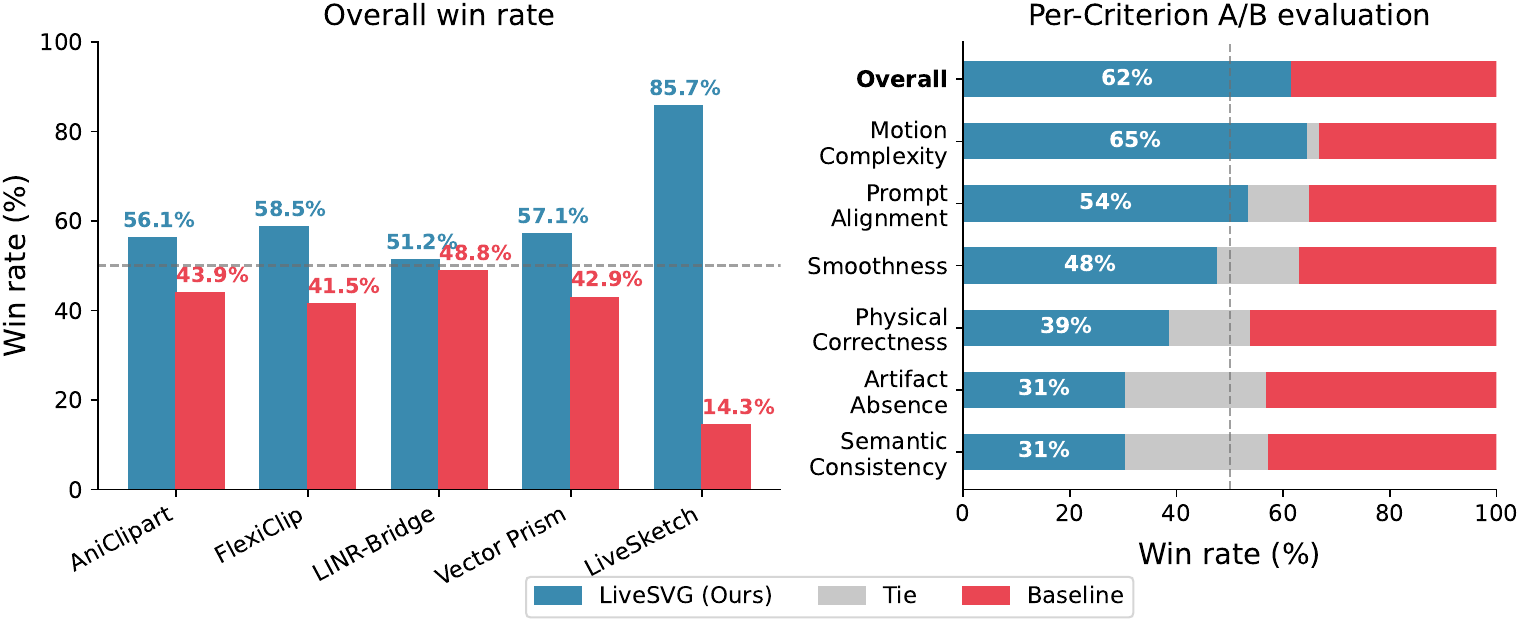}
  \caption{Automated A/B preference study results for \emph{\ours (LTX-Video)} with an LLM-based judge.
           Left: per-criterion outcomes.
           Right: overall win rate against each baseline.
           The LTX-Video variant wins 61.7\% of 209 overall comparisons.}
  \Description{Two charts summarizing the A/B study for the LiveSVG LTX-Video variant. The left
  panel shows per-criterion win, tie, and baseline-win percentages. The right panel shows
  overall win rates against AniClipart, FlexiClip, LINR-Bridge, Vector Prism, and LiveSketch.}
  \label{fig:ab_results_ltx}
\end{figure*}

\paragraph{AniClipart automated A/B results.}
Figure~\ref{fig:ab_results} reports the automated A/B preference study for the main
AniClipart setting.
\Cref{fig:ab_results_ltx} reports the same protocol for the \emph{\ours (LTX-Video)}
variant.

\section{Ablation Study}
\label{sec:appendix_ablation_study}

\paragraph{Automated ablations.}
We ablate four components of \ours using Gemini~3.1~Pro~\cite{team2024gemini} A/B tests
(\Cref{tab:appendix_ablation_ab}).
The ablations remove per-group homography learning, tracking-point initialization, exponential
spatial regularization, or $G^1$ regularization.
The full method wins against all reported ablations.

\input{tables/ablation_ab}

\begin{figure}[t]
  \centering
  \includegraphics[width=0.82\linewidth]{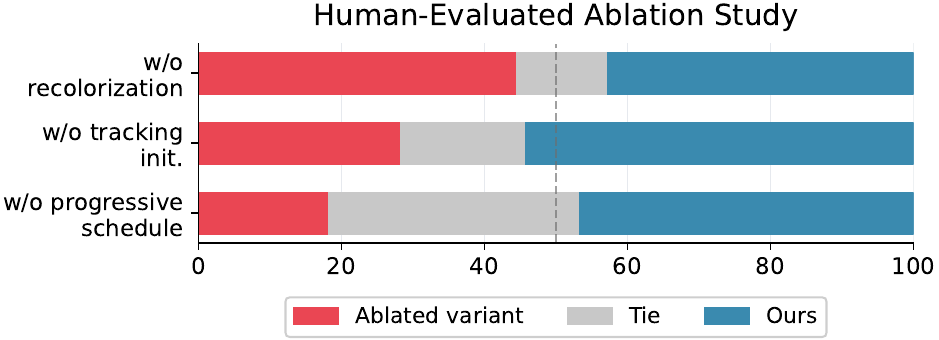}
  \caption{Human artifact ablation study on AniClipart.
           Workers choose which animation has fewer visible artifacts.
           Bars report the fraction of votes for the full method, ties, and the corresponding
           ablated variant, with duplicated consistency trials removed and workers exceeding the
           duplicate-trial inconsistency threshold filtered.}
  \Description{Horizontal stacked bar chart for three ablations: no sphere recolorization, no
  tracking initialization, and no progressive schedule. The no-recolorization comparison is nearly
  tied, while the full method receives more fewer-artifact votes than the no-tracking and
  no-progressive-schedule variants.}
  \label{fig:appendix_ablation_user_study}
\end{figure}

\paragraph{Human artifact ablations.}
We additionally conduct a targeted AMT ablation study on AniClipart that isolates visible artifacts rather than overall preference.
Each comparison shows the same reference SVG and two anonymized animations, and workers choose which animation contains fewer visible artifacts, with a tie option. We compare the full method against three variants: no sphere-packing recolorization, no tracking-based progressive initialization, and no progressive schedule.
\Cref{fig:appendix_ablation_user_study} summarizes the results after applying the same duplicate
consistency-control filtering used in the main human study.

For the no-recolorization ablation, workers do not favor either method: the full method receives
42.7\% of fewer-artifact votes, the original-color variant receives 44.5\%, and 12.7\% are ties. This suggests that sphere-packing recolorization does not introduce a systematic artifact penalty.
This comparison is less controlled than the other two ablations because the video model receives different input colors in the two conditions, and therefore generates different target motions. Although workers are instructed to evaluate artifacts only, complex motions can naturally be more artifact-prone. We therefore interpret this experiment as evidence that recolorization does not harm the final animation quality, rather than as a strict fixed-target comparison. In practice, recolorization remains important when similarly colored elements cross or overlap,
where fixed-color pixel supervision can otherwise produce ambiguous correspondence signals.

For the tracking and schedule ablations, the comparisons are more directly diagnostic of the SVG optimization pipeline. The full method is judged to have fewer artifacts more often than the no-tracking-initialization variant (54.2\% vs. 28.3\%, with 17.5\% ties) and the no-progressive-schedule variant (46.7\% vs. 18.3\%, with 35.0\% ties). These results support both components as integral to stable optimization: tracking-based initialization improves alignment when new frames are activated, while the progressive schedule avoids optimizing distant poses from weak initial overlap.

\section{Evaluation Protocol}

\paragraph{Dataset and baselines.}
Experiments are conducted on the AniClipart benchmark (43 examples) and the \ourhard
benchmark introduced in the main paper, using the same SVG and text prompt per example for all
methods.
We compare against five baselines: AniClipart, FlexiClip, LiveSketch, LINR-Bridge, and
Vector Prism.
Baselines are evaluated using the authors' available implementations or released outputs when
available, and all animations are rendered to a common raster representation before computing
metrics.

\paragraph{Runtime comparison.}
We also benchmark runtime on five common AniClipart samples using
\texttt{Evaluation/analyze\_runtime.py}.
The timing call starts immediately before each method subprocess and stops after it exits.
Consequently, the reported wall-clock time includes all initialization inside each implementation,
including model loading, preprocessing, skeleton or keypoint extraction, triangulation setup,
optimization initialization, and export when those stages are used by the method.
This is important for skeleton-based baselines, where the initialization cost is part of the
practical latency paid by a user rather than a separate offline quantity.
GPU methods are run with exclusive access to one A100 GPU and with the same 100-iteration
benchmark budget, so the mean time and mean seconds per iteration in
\Cref{tab:appendix_runtime} provide an iteration-normalized comparison.
Vector Prism is LLM-based and has no gradient-iteration count; for it, we report wall-clock time,
API latency-dominated runtime, average API calls, and token usage.
Under this protocol, \ours is the fastest GPU optimization method at 5.2 minutes per successful
sample, compared with 6.9 minutes for AniClipart, 10.7 minutes for LINR-Bridge, 14.5 minutes for
FlexiClip, and 22.3 minutes for LiveSketch.

\input{tables/runtime_comparison}

\paragraph{Frame normalization and metric computation.}
To compare methods with different output formats, including animated SVG, GIF, MP4, and
HTML/CSS animation,
we render all outputs to raster frames under a common protocol.
Motion-aware metrics are computed after resampling each animation to 24 frames.
For XCLIP and cross-seed Diversity, we uniformly sample 8 frames per video, encode them with
X-CLIP video embeddings, and use cosine-based similarity/distance. LPIPS and SSIM are computed against the static reference rendering and are therefore reported
together with motion-sensitive metrics.

\paragraph{Automated A/B judging.}
As a complementary evaluation signal, we run automated pairwise preference tests with an
LLM-based judge, summarized in the main paper.
Each comparison shows the same reference SVG rendering, the same text prompt, and two unlabeled
MP4 animations in randomized order.
The judge reports criterion-level winners and an overall winner across seven criteria
(Smoothness, Motion Complexity, Artifact Absence, Prompt Alignment, Semantic Consistency,
Physical Correctness, Overall Satisfaction).
The main AniClipart setting includes 206 comparisons, the LTX-Video variant includes
209 comparisons, and \ourhard includes 99 comparisons.

The judge prompt instructs Gemini to act as an expert animation reviewer. It receives the static
SVG rendering, Animation A, Animation B, and the intended caption. For each criterion, it must
provide a one- or two-sentence justification and choose A, B, or Tie; it then provides an overall
justification and selects a forced overall winner. The prompt explicitly states that animations
with meaningful complex motion should generally be favored over nearly static animations, provided
that visual artifacts remain minor.

\section{Human Preference Study}
\label{sec:appendix_user_study}

We conducted two human user studies on Amazon Mechanical Turk (AMT): one for the AniClipart benchmark and one for \ourhard. Both studies used the same pairwise A/B interface and the same quality-control protocol. The study measured forced overall preference between two unlabeled animations generated for the same static SVG reference image and motion prompt.

\paragraph{Worker eligibility.}
We restricted participation to workers satisfying all of the following AMT qualifications:
Mechanical Turk Masters status; HIT approval rate of at least 98\%; at least 500 approved HITs;
and location in one of the following English-speaking countries: United States, United Kingdom,
Canada, Australia, or India.

\begin{figure*}[t]
  \centering
  \includegraphics[width=0.92\linewidth]{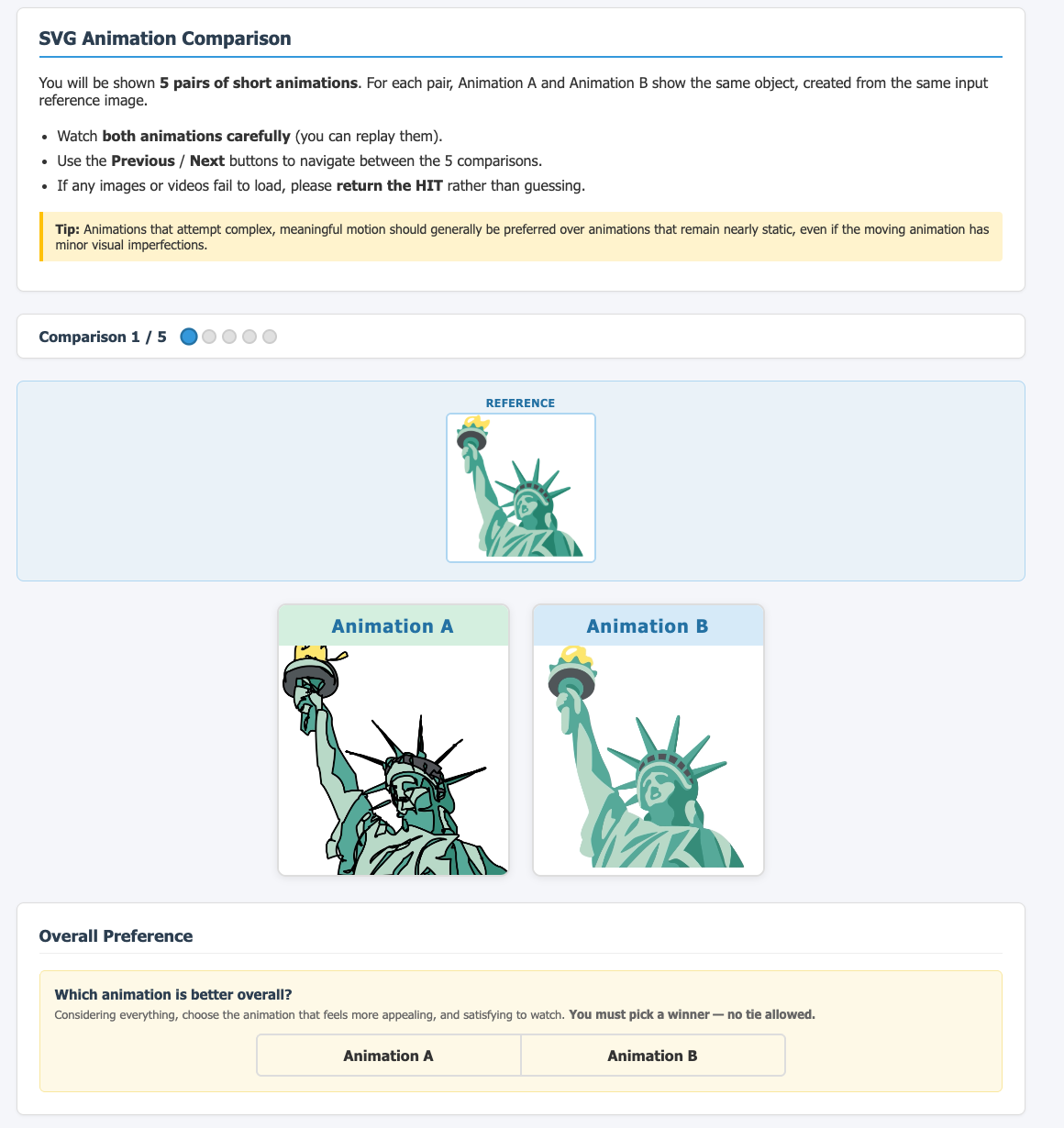}
  \caption{Amazon Mechanical Turk interface for the human A/B preference studies.
           Workers viewed the static SVG reference and two anonymized animations, then selected
           the animation they preferred overall.}
  \Description{Screenshot of an AMT pairwise animation-comparison interface with instructions,
  a progress indicator, a reference image, two side-by-side animations labeled Animation A and
  Animation B, and a forced overall preference question.}
  \label{fig:amt_screenshot}
\end{figure*}

\paragraph{Task interface and instructions.}
Each comparison page showed the rendered static SVG reference above two MP4 animations labeled
only as ``Animation A'' and ``Animation B.'' Videos were displayed side by side, autoplayed,
looped, muted, and could be replayed by the worker. Workers navigated through the batch with
\emph{Previous} and \emph{Next} buttons, and the interface prevented advancing or submitting
until the current comparison had been answered. The exact worker-facing instructions were:
\begin{quote}\small\itshape
\textbf{SVG Animation Comparison.}
You will be shown a batch of pairs of short animations. For each pair, Animation A and
Animation B show the same object, created from the same input reference image.

Watch both animations carefully (you can replay them).
Use the Previous / Next buttons to navigate between the comparisons.
If any images or videos fail to load, please return the HIT rather than guessing.

\textbf{Tip:} Animations that attempt complex, meaningful motion should generally be preferred
over animations that remain nearly static, even if the moving animation has minor visual
imperfections.

\textbf{Overall Preference.}
Which animation is better overall?
Considering everything, choose the animation that feels more appealing and satisfying to watch.
You must pick a winner---no tie allowed.
\end{quote}
The production HITs instantiated the same template with either 5 or 10 comparisons per worker
batch, depending on the batch size used for that round of collection. Figure~\ref{fig:amt_screenshot}
shows the AMT interface used in the studies.

\paragraph{Randomization and blinding.}
For every pairwise comparison, the assignment of methods to A/B positions was randomized
independently. All image and video assets were served through hashed file paths, so visible URLs
and asset names did not reveal which method produced either animation. The displayed interface
contained only the reference image, the two anonymized animations, and the overall-preference
question; method identities were stored only in hidden metadata fields for later aggregation.

\paragraph{AniClipart study configuration.}
For AniClipart, each baseline was compared against three variants of \ours, corresponding to the
three reference-video generators evaluated in the paper: \emph{\ours (Veo~3.1)},
\emph{\ours (LTX~2.3)}, and \emph{\ours (WAN~2.2)}. Thus, the AniClipart human study collected
preferences for each \ours video-model variant separately, in addition to the per-example
best-of-three aggregation reported in the main paper.

\paragraph{\ourhard study configuration.}
For \ourhard, we tested only the \emph{\ours (WAN~2.2)} variant. Each hard-benchmark comparison
paired this variant against one baseline output for the same SVG and prompt, using the same
randomized and blinded A/B interface as the AniClipart study.

\paragraph{Consistency controls and filtering.}
Each worker batch contained two duplicated comparisons as attention and consistency controls.
The duplicated comparisons used the same method and example, but were still shown through the
same randomized A/B assignment mechanism. During post-processing, we compared the worker's
answers on each duplicated pair after mapping A/B choices back to method identities.
Workers with more than 50\% inconsistent answers on these repeated comparisons were filtered out
automatically. For retained workers, duplicated control comparisons were used for quality control
and only one copy of each duplicated comparison was counted in the final preference totals.

\section{Implementation Details}

\paragraph{Preprocessing.}
Given an input SVG and a text prompt, \ours first uses Gemini~3.1~Pro~\cite{team2024gemini} for
zero-shot semantic grouping and caption refinement.
The model receives the SVG source together with a raster rendering and is asked to insert flat,
non-nested semantic groups while preserving the original drawing order when possible.
The paths in the grouped SVG are then recolored with a high-separation palette
(\Cref{sec:appendix_colorization}) so that pixel-space supervision can distinguish parts with
similar original colors.
Layer order is corrected when the reference motion indicates inconsistent occlusions, as described
in \Cref{sec:appendix_opt_layers}.

\paragraph{Reference-video generation.}
The main variant uses Veo~3.1~\cite{brooks2024veo} as a frozen image-to-video model.
We also evaluate Wan Video~\cite{wan2025} and LTX-Video~\cite{hacohen2025ltxvideo} variants.
For each input, the recolored SVG is rendered to an image and passed to the video model with the
following prompt template:
\begin{quote}\small\itshape
Animate this image as flat 2D vector-style animation of [motion prompt].
Style: pure 2D motion graphics, solid block colors, crisp edges, no gradients, no shading, and
zero 3D depth.
Motion: restrict all movement strictly to the 2D image plane using simple articulated or
puppet-style motion.
Consistency: preserve the original geometric shapes and exact colors; maintain the object's
current facing direction; do not add new details or remove existing details.
Environment: solid white background with no new scene content.
\end{quote}
For each SVG, we generate 10--20 candidate videos using different random seeds.
The selected candidate is not chosen by visual appeal alone; it must be suitable as a fixed-target
sequence for optimizing an SVG with fixed topology and fixed colors.

\paragraph{Gemini scoring and reranking.}
Candidate videos are selected with a two-stage Gemini~3.1~Pro procedure.
In the first stage, Gemini receives a short MP4 clip and representative keyframes from one
candidate video. The first frame serves as the reference for color and element preservation.
Gemini is instructed to evaluate whether the video can serve as a ground-truth target for
fixed-topology SVG fitting by pixel-space MSE. The prompt template is:
\begin{quote}\small\itshape
You are an expert evaluator for 2D vector-style animations. You will be shown a short video
generated from a static SVG image. The goal is to determine whether this video can serve as a
ground-truth target for downstream SVG optimization. In this optimization, a fixed set of SVG
shapes tracks the video frame by frame by minimizing pixel-space MSE. The shapes can deform, but
their fill colors are fixed and no new shapes can be added.

Evaluate the video according to four constraints: (1) no 3D rotation that reveals hidden
geometry; 2D stretching, morphing, bending, and squashing are allowed; (2) no new semantic
elements beyond those present in the first frame; (3) constant flat colors with no lighting,
shadows, highlights, color grading, or brightness shifts; and (4) physical continuity, meaning
that elements should not suddenly appear from nowhere or dissolve, although ordinary occlusion is
allowed.

For each constraint, provide a short analysis and a categorical score: Yes, Partial, or No.
Finally, decide whether the video is viable as an optimization target for fixed-color SVG
tracking.
\end{quote}
The first-stage response contains analyses of motion/deformation, invented details, physical
continuity, color constancy, and overall suitability.
The top candidates then enter a second comparative stage. Gemini receives the candidate clips,
the same representative keyframes for each candidate, and the following instruction:
\begin{quote}\small\itshape
You are an expert evaluator for 2D vector-style animations. You will be shown multiple candidate
videos generated from the same static SVG image. Compare them and rank them by suitability as
ground-truth targets for downstream SVG optimization.

A better target keeps all motion strictly 2D, preserves only the original elements, maintains
constant flat fill colors, and preserves physical continuity. When comparing candidates, favor
the one with the least color drift, cleanest 2D motion, fewest invented elements, and most stable
structure. A slightly less dynamic but clean video is preferred over a dramatic video with
artifacts, because consistency is more important than expressiveness for pixel-level SVG fitting.

Analyze each candidate, discuss trade-offs, and return a ranking from best to worst.
\end{quote}
The highest-ranked candidate is used as the fixed reference sequence for SVG fitting.

\paragraph{Differentiable SVG fitting.}
We optimize the original SVG geometry with DiffVG~\cite{li2020diffvg}.
Each semantic group has an absolute full 8-DOF homography at every keyframe, mapping the original
undeformed group directly to its pose in that keyframe.
This avoids accumulated error from composing frame-to-frame transforms.
Each path additionally has keyframe-dependent learnable offsets for its B\'ezier control points,
allowing local non-rigid deformation beyond the group transform.
The data term is a foreground-cleaned, Gaussian-smoothed pixel MSE loss between rendered SVG
frames and selected reference keyframes.
Before computing the loss, target-frame backgrounds are cleaned with a slightly dilated foreground
mask so that near-white video-model artifacts do not create spurious gradients.
Local path offsets are regularized by exponential spatial smoothness, $G^1$ tangent continuity,
and foreground SDF containment.

\begin{table}[t]
  \centering
  \small
  \caption{Main experimental configuration for \ours.}
  \begin{tabular}{p{0.36\linewidth}p{0.56\linewidth}}
    \toprule
    Component & Setting \\
    \midrule
    Reference generator & Veo~3.1 by default; Wan Video and LTX-Video for reference-generator variants \\
    Candidate generation & 10--20 videos per input, sampled with different random seeds \\
    Candidate selection & Gemini two-stage scoring and comparative reranking using video clips and representative keyframes \\
    Optimization frames & 15 keyframes, also used as the exported animation frames \\
    Optimization budget & 2000 iterations \\
    Renderer & DiffVG differentiable rasterization \\
    Resolution & $256\times256$ optimization; 720-pixel visualizations \\
    Global motion & Absolute full 8-DOF homography per semantic group and keyframe \\
    Local motion & Per-path, per-keyframe B\'ezier control-point offsets \\
    Optimizer & Adam, $\beta=(0.9,0.9)$, $\epsilon=10^{-6}$ \\
    Learning rates & $10^{-3}$ for homographies; $10^{-1}$ for path offsets \\
    Data term & Foreground-cleaned, Gaussian-smoothed MSE with weight 1000 \\
    Regularization & Spatial smoothness, $G^1$ continuity, and foreground SDF containment \\
    Regularization weights & 0.5, 10.0, and 1.0, respectively \\
    Foreground margin & One-pixel tolerance for background cleanup and SDF containment \\
    Progressive schedule & Progressive activation of 15 keyframes, approximately one new keyframe every 100 iterations, followed by joint refinement \\
    \bottomrule
  \end{tabular}
  \label{tab:appendix_default_setup}
\end{table}

\paragraph{Progressive schedule.}
We use progressive frame optimization. All keyframes are initialized from the static SVG, but only
the early frames are optimized first. Then, approximately every 100 iterations, one additional
keyframe is activated. The new keyframe copies the homography and path-deformation parameters from
the preceding active keyframe and is further initialized with the tracking procedure in
\Cref{sec:appendix_tracking_init}. Once all 15 keyframes are active, the remaining iterations
jointly refine the full sequence. This schedule improves initial overlap at each frame and avoids
weak gradients that arise when distant poses are optimized directly from the rest pose.

\paragraph{Output representation.}
The final artifact is a standard animated SVG with keyframed transforms and keyframed path
geometry. Recolorization is used only for reference generation and supervision; exported
animations are restored to the original colors.

\section{Models Used Across Pipeline and Evaluation}

Table~\ref{tab:appendix_models} summarizes the main external models and toolkits used in
preprocessing, optimization, and evaluation.

\begin{table}[t]
  \centering
  \small
  \caption{External models and toolkits used in \ours.}
  \begin{tabular}{p{0.35\linewidth}p{0.57\linewidth}}
    \toprule
    Component & Model / toolkit \\
    \midrule
    Semantic grouping + captioning & Gemini~3.1~Pro~\cite{team2024gemini} \\
    Reference video generation & Veo~3.1~\cite{brooks2024veo} (default), Wan Video~\cite{wan2025} and LTX-Video~\cite{hacohen2025ltxvideo} (variants) \\
    Differentiable renderer & DiffVG~\cite{li2020diffvg} \\
    Tracking-based keyframe initialization & TAPNext~\cite{doersch2024tapnext} \\
    Layer reordering & SAM2~\cite{ravi2024sam2} \\
    Foreground masking & RMBG-1.4 foreground segmentation \\
    \bottomrule
  \end{tabular}
  \label{tab:appendix_models}
\end{table}

\section{Additional Module Details}

\paragraph{Implementation details.}
The local deformation and group motion are composed as path offsets followed by an
absolute group homography: for original control point $x_i^0$, global SVG center
$c_0$, center-relative point $u_i=x_i^0-c_0$, keyframe offset $\Delta_{k,i}$, and
group $g(i)$, the rendered point is
$\hat{x}_{k,i}=\pi(H_{g(i),k}[c_0+u_i+\Delta_{k,i},1]^\top)$, where
$\pi([x,y,w]^\top)=[x/w,y/w]^\top$ and the full homography is composed as
$H_{g,k}=P\,T\,T_{c_g}\,R\,Sh\,S\,T_{-c_g}$. This center shift allows the shape to rotate around itself (around a learned center).
The pixel loss applies Gaussian-smoothed MSE to both rendered and reference frames
using a $5\times5$ gaussian kernel with $\sigma=1.0$.
Foreground masks are extracted by thresholding the RMBG-1.4 foreground alpha at
$0.5$. For target cleanup the mask is dilated by one pixel and pixels outside it
are whitened, while the SDF term uses $D_k(x)=\max(0,\operatorname{EDT}(1-M_k)(x)-1)$ and samples this map at raw control-point locations. Exponential spatial regularization weights adjacent control-point offset 
differences by $w_{ij}=\exp[-(\|x_i^0-x_j^0\|_2/\sigma_s)^2]$ with
$\sigma_s=0.01W$, where $W$ is the SVG canvas width.
For cubic B\'ezier joints $(c^-_j,a_j,c^+_j)$, the $G^1$ loss is
$\operatorname{mean}_{k,j}\left(1-\cos(a_j-c^-_j,c^+_j-a_j)\right)$, applied to
joints that are smooth in the rest SVG up to the $10^\circ$ enforcement threshold (waiving sharp corners). TAPNext initialization, tracks explicit on-curve anchors from
earlier rendered keyframes to the target keyframe, accepts visible predictions with
certainty at least $0.9$, maps selected target positions through the inverse group
homography to pre-homography coordinates, and writes the resulting updates to the
local path-offset parameters; off-curve handles are either co-translated or updated
by the segment-chord similarity rule when that option is enabled.

\subsubsection{Automatic Layer Reordering}
\label{sec:appendix_opt_layers}

Incorrect painter's order can make otherwise accurate geometry render implausibly.
Our layer module estimates a better order before optimization.

\paragraph{Tracked-mask evidence.}
For each semantic group, we render an isolated binary mask at frame 0 and use it to initialize
SAM2 propagation over the reference sequence.
Let $A_i^0$ denote the baseline area of group $i$, and let $A_i^{\mathrm{exc}}(t)$ be its
exclusive tracked area at frame $t$ (pixels jointly claimed by multiple groups are ignored).
We compute
\[
R_i(t) = \frac{A_i^{\mathrm{exc}}(t)}{A_i^0}.
\]
For overlapping group pairs $(i,j)$, we define a weighted score
\[
s_{i,j} =
\frac{\sum_t w_{i,j}(t)\big(R_i(t)-R_j(t)\big)}
     {\sum_t w_{i,j}(t)},
\]
where $w_{i,j}(t)$ is the intersection area of their tracked bounding boxes at frame $t$.

\paragraph{Order recovery and safeguards.}
If $|s_{i,j}|$ is below a small threshold, the pair is treated as ambiguous.
Otherwise, the sign determines foreground preference and induces a directed edge in a group-order
graph.
We recover a global order via topological sorting, with in-degree ties resolved by original SVG
order.
If the graph is cyclic or if a proposed reorder changes rendering unexpectedly, we fall back
conservatively to incremental swaps and, when needed, local path-level refinement.

\subsubsection{Sphere-Packing Recolorization}
\label{sec:appendix_colorization}

If two paths have similar colors, pixel MSE gives ambiguous correspondence signals.
We therefore assign $K$ paths a high-separation palette of $K$ RGB colors from
precomputed best-known sphere packings in $[0,1]^3$ (Packomania~\cite{packomania}).
Colors are assigned after a deterministic shuffle, prioritizing larger paths first.
Recolorization is applied consistently to both SVG renders and reference frames during fitting,
then removed in the final export.
If a packing for a specific $K$ is unavailable, we use an HSV distinct-color fallback.

\subsubsection{Target Cleanup and Path Regularization}

The reference videos sometimes contain faint off-white background artifacts even when the prompt
asks for flat vector colors.
Before computing the pixel reconstruction loss, we segment the foreground in each target keyframe,
dilate the mask by one pixel, and set all pixels outside the padded mask to pure white.
The rendered SVG frame itself is not masked; only the target is cleaned.

For path regularization, the main configuration combines three losses with weights
0.5, 10.0, and 1.0, respectively.
The spatial term penalizes inconsistent displacement between neighboring control points, with an
exponential falloff based on their original distance.
The $G^1$ term penalizes tangent-direction discontinuities at cubic B\'ezier joints to suppress
visible kinks.
The SDF term constructs a per-keyframe Euclidean distance field from the foreground mask and
penalizes raw control points that move more than one pixel outside the foreground support.

\subsubsection{Tracking-Based Progressive Initialization}
\label{sec:appendix_tracking_init}

At each progressive transition, copying parameters from the previous keyframe can be insufficient
under large motion changes.
We sample explicit on-curve anchors from the current SVG render and track them to
the incoming reference frame using TAPNext.
High-certainty predictions are mapped back through the inverse group homography
and used to initialize the new keyframe's local path offsets.
Off-curve tangent handles are propagated from the tracked anchors, either by
co-translation or by the local similarity induced by the tracked segment chord.
This module is a warm start only; all parameters are still optimized jointly
afterward.

%% file: tables/benchmark_svg_stats.tex
\begin{table*}[t]
  \centering
  \small
  \caption{Structural statistics of the SVG benchmarks. ``Elements'' counts non-root SVG XML elements excluding metadata, title, and description nodes. ``Colors'' counts distinct explicit fill/stroke/stop colors. ``Path coord. nums.'' counts numeric coordinate values in path \texttt{d} attributes, which serves as a proxy for the amount of B\'ezier/path geometry.}
  \begin{tabular}{lrrrrrrrrrr}
    \toprule
    Benchmark & \#SVGs & \multicolumn{3}{c}{Elements} & \multicolumn{3}{c}{Colors} & \multicolumn{3}{c}{Path coord. nums.} \\
    \cmidrule(lr){3-5} \cmidrule(lr){6-8} \cmidrule(lr){9-11}
     & & Mean & Med. & Max & Mean & Med. & Max & Mean & Med. & Max \\
    \midrule
    AniClipart & 43 & 26.7 & 23 & 60 & 6.5 & 6 & 15 & 1149.4 & 946 & 3162 \\
    \ourhard & 35 & 54.9 & 45 & 140 & 17.5 & 18 & 37 & 1415.5 & 1337 & 2497 \\
    \bottomrule
  \end{tabular}
  \label{tab:benchmark_svg_stats}
\end{table*}

%% file: tables/metrics_svgx_hard.tex
\begin{table}[t]
  \caption{Quantitative comparison on the \ourhard benchmark.
           For optimization-based methods, \textbf{bold} marks the best value and
           underlining marks the second best; ties share the same emphasis.
           The VLM baseline generates animation programs directly; in our benchmark outputs,
           these programs are usually transform-centric and only rarely animate path geometry
           explicitly.
           Optimization-based methods instead update continuous motion parameters, making
           articulated geometric deformation a first-class part of the representation.
           DOVER~\cite{wu2023dover} measures overall perceptual video quality.
           LTX and WAN denote the corresponding reference-video generator variants.
           Time/SVG reports mean wall-clock minutes per successful sample under the runtime
           protocol in \Cref{tab:appendix_runtime_svgx_hard}.
           VMem reports mean net GPU memory usage in GB under the same protocol.}
  \label{tab:quant_svgx_hard}
  \centering
  \small
  \setlength{\tabcolsep}{5pt}
  \resizebox{\columnwidth}{!}{%
\begin{tabular}{l cccccc}
  \toprule
  Method & XCLIP$\uparrow$ & LPIPS$\downarrow$ & SSIM$\uparrow$ & DOVER$\uparrow$ & Time/SVG$\downarrow$ & VMem (GB)$\downarrow$ \\
  \midrule
  No Animation & 0.214 & \textbf{0.000} & \textbf{1.000} & 0.470 & -- & -- \\
  \midrule
  \multicolumn{7}{l}{\textit{LLM/VLM-based}} \\[2pt]
  Vector Prism & 0.211 & 0.139 & 0.867 & 0.461 & 4.4 & -- \\
  \midrule
  \multicolumn{7}{l}{\textit{Optimization based on video SDS loss}} \\[2pt]
  LiveSketch & 0.182 & 0.503 & 0.609 & 0.397 & 23.6 & 39.0 \\
  AniClipart & 0.204 & \underline{0.274} & \underline{0.781} & \underline{0.433} & 87.4 & 28.1 \\
  FlexiClip & 0.201 & 0.298 & 0.773 & 0.424 & 219.3 & 28.1 \\
  LINR-Bridge & \underline{0.205} & 0.491 & 0.729 & 0.426 & 142.1 & 17.1 \\
  \midrule
  \multicolumn{7}{l}{\textit{Target (mp4) video based}} \\[2pt]
  \textbf{\ours (WAN~2.2)} & \textbf{0.215} & \textbf{0.208} & \textbf{0.844} & \textbf{0.476} & \textbf{10.2} & \textbf{7.3} \\
  \bottomrule
\end{tabular}
  }
\end{table}

%% file: tables/runtime_comparison_svgx_hard.tex
\begin{table}[t]
  \centering
  \small
  \setlength{\tabcolsep}{4pt}
  \caption{Runtime comparison on ten \ourhard samples. GPU methods were run for 50 optimization iterations with exclusive access to one NVIDIA A100-SXM4-80GB GPU; successful samples only are aggregated. Timing wraps the complete method subprocess, so initialization, model loading, preprocessing, skeleton/keypoint estimation, triangulation setup, optimization, and export are included when present. Vector Prism is reported separately as an LLM/API baseline with no gradient-iteration count.}
  \label{tab:appendix_runtime_svgx_hard}
  \resizebox{\linewidth}{!}{%
  \begin{tabular}{llccccc}
    \toprule
    Method &
    Type &
    \shortstack{Rec.\\iters} &
    \shortstack{Bench\\iters} &
    \shortstack{Mean\\min/SVG} &
    \shortstack{Median\\min/SVG} &
    \shortstack{Mean\\sec/iter} \\
    \midrule
    Vector Prism & LLM code synthesis & -- & -- & 4.4 & 3.7 & -- \\
    \midrule
    AniClipart & Video-SDS optimization & 500 & 50 & 17.5 & 15.0 & 21.02 \\
    FlexiClip & Video-SDS optimization & 500 & 50 & 31.4 & 28.1 & 37.70 \\
    LINR-Bridge & Video-SDS optimization & 10000 & 50 & 14.4 & 14.3 & 17.31 \\
    LiveSketch & Video-SDS optimization & 500 & 50 & 2.7 & 2.5 & 3.22 \\
    \ours & Reference-video fitting & 1500 & 50 & 1.5 & 1.6 & 1.84 \\
    \bottomrule
  \end{tabular}
  }
\end{table}

%% file: tables/ablation_ab.tex
\begin{table}[t]
  \centering
  \small
  \caption{Ablation study with Gemini~3.1~Pro A/B judging.
           Each row compares the full \ours pipeline against a variant with one component removed
           and reports the judge's overall decision.}
  \label{tab:appendix_ablation_ab}
  \setlength{\tabcolsep}{4pt}
  \resizebox{\columnwidth}{!}{%
  \begin{tabular}{l l c}
    \toprule
    Variant & Removed component & \ours overall wins \\
    \midrule
    w/o homography & Per-group homography learning & 22/42 (52.4\%) \\
    w/o tracking init. & Tracking-point initialization & 28/42 (66.7\%) \\
    w/o spatial reg. & Exponential spatial regularization & 30/42 (71.4\%) \\
    w/o $G^1$ reg. & $G^1$ continuity regularization & 22/42 (52.4\%) \\
    \bottomrule
  \end{tabular}
  }
\end{table}

%% file: tables/runtime_comparison.tex
\begin{table}[t]
  \centering
  \small
  \setlength{\tabcolsep}{4pt}
  \caption{Runtime comparison on ten AniClipart samples. GPU methods were run for 50 optimization iterations with exclusive access to one NVIDIA A100-SXM4-80GB GPU; successful samples only are aggregated. Timing wraps the complete method subprocess, so method-specific initialization, model loading, preprocessing, skeleton/keypoint estimation, triangulation setup, optimization, and export are included when present. Vector Prism is reported separately as an LLM/API baseline with no gradient-iteration count.}
  \label{tab:appendix_runtime}
  \resizebox{\linewidth}{!}{%
  \begin{tabular}{llccccc}
    \toprule
    Method &
    Type &
    \shortstack{Rec.\\iters} &
    \shortstack{Bench\\iters} &
    \shortstack{Mean\\min/SVG} &
    \shortstack{Median\\min/SVG} &
    \shortstack{Mean\\sec/iter} \\
    \midrule
    Vector Prism & LLM code synthesis & -- & -- & 3.1 & 3.1 & -- \\
    \midrule
    AniClipart & Video-SDS optimization & 500 & 50 & 3.2 & 3.2 & 3.88 \\
    FlexiClip & Video-SDS optimization & 500 & 50 & 8.0 & 7.9 & 9.63 \\
    LINR-Bridge & Video-SDS optimization & 10000 & 50 & 8.1 & 7.3 & 9.77 \\
    LiveSketch & Video-SDS optimization & 500 & 50 & 4.1 & 3.9 & 4.88 \\
    \ours & Reference-video fitting & 1500 & 50 & 1.4 & 1.4 & 1.66 \\
    \bottomrule
  \end{tabular}
  }
\end{table}